\documentclass[10pt,journal,compsoc]{IEEEtran}
\usepackage{multirow}
\usepackage{booktabs}
\usepackage{xcolor}
\usepackage{url}
\usepackage{mdframed}
\usepackage{graphicx}
\usepackage{amsmath,amsfonts}
\usepackage{algorithmic}
\usepackage{algorithm}
\usepackage{colortbl}

\usepackage{subcaption}
\usepackage{tikz}
\usetikzlibrary{positioning, arrows.meta, decorations.pathreplacing, calc}
\newcommand*{\circled}[1]{\textcircled{\raisebox{-.9pt}{#1}}}

\newcommand{\E}{\mathbb{E}}
\newcommand{\I}{\mathcal{I}} % Mutual Information
\newcommand{\Hc}{\mathcal{H}} % Entropy

\ifCLASSOPTIONcompsoc
  \usepackage[nocompress]{cite}
\else
  \usepackage{cite}
\fi
\ifCLASSINFOpdf
\else
\fi
\begin{document}
\title{CoIFNet: A Unified Framework for Multivariate Time Series Forecasting with Missing Values}

\author{
    \IEEEauthorblockN{
        Kai~Tang,
        Ji~Zhang*,
        Hua Meng,
        Minbo~Ma,
        Qi~Xiong,
        Fengmao Lv,
        Jie~Xu,
        Tianrui~Li
    } \\
    \IEEEcompsocitemizethanks{
        \IEEEcompsocthanksitem Kai~Tang, Ji~Zhang, Qi~Xiong, Fengmao Lv and Tianrui~Li are with the School of Computing and Artificial Intelligence, Southwest Jiaotong University, Chengdu, China. E-mail: jizhang.jim@gmail.com. *Ji Zhang is the corresponding author.
        \IEEEcompsocthanksitem Hua~Meng is with the School of Mathematics, Southwest Jiaotong University, Chengdu, China. Email: menghua@swjtu.edu.cn.
        \IEEEcompsocthanksitem Minbo~Ma is with the Institute Carbon Neutrality, Tsinghua University, P.R. China. Email: minbo46.ma@gmail.com.
        \IEEEcompsocthanksitem Jie~Xu is with the White Rose Grid e-Science Centre, University of Leeds, England. Email: J.Xu@leeds.ac.uk.
        % \IEEEcompsocthanksitem 
    }
    % \IEEEauthorblockA{School of Computing and Artificial Intelligence, Southwest Jiaotong University, Chengdu, China} \\
    % \IEEEauthorblockA{\{tangkai, minboma, xiongqi\}@my.swjtu.edu.cn, \{jizhang, trli\}@swjtu.edu.cn}
    % \thanks{Manuscript received April 19, 2021; revised August 16, 2021. This work was supported in part by}

}


\markboth{Journal of \LaTeX\ Class Files,~Vol.~14, No.~8, August~2015}%
{Shell \MakeLowercase{\textit{et al.}}: Bare Advanced Demo of IEEEtran.cls for IEEE Computer Society Journals}



\IEEEtitleabstractindextext{%
    \begin{abstract}
        Multivariate time series forecasting (MTSF) is a critical task with broad applications in domains such as meteorology, transportation, and economics. Nevertheless, pervasive missing values caused by sensor failures or human errors significantly degrade forecasting accuracy. Prior efforts usually employ an impute-then-forecast paradigm, leading to suboptimal predictions due to error accumulation and misaligned objectives between the two stages.
        To address this challenge, we propose the Collaborative Imputation-Forecasting Network (CoIFNet), a novel framework that unifies imputation and forecasting to achieve robust MTSF in the presence of missing values.
        Specifically, CoIFNet takes the observed values, mask matrix and timestamp embeddings as input, processing them sequentially through the Cross-Timestep Fusion (CTF) and Cross-Variate Fusion (CVF) modules to capture temporal dependencies that are robust to missing values. We provide theoretical justifications on how our CoIFNet learning objective improves the performance bound of MTSF with missing values.
        Through extensive experiments on challenging MSTF benchmarks, we demonstrate the effectiveness and computational efficiency of our proposed approach across diverse missing-data scenarios, e.g., CoIFNet outperforms the state-of-the-art method by \underline{\textbf{24.40}}\% (\underline{\textbf{23.81}}\%) at a point (block) missing rate of 0.6,
        while improving memory and time efficiency by  $\underline{\boldsymbol{4.3\times}}$ and $\underline{\boldsymbol{2.1\times}}$, respectively.
        Our code is available at: \textcolor{magenta}{\url{https://github.com/KaiTang-eng/CoIFNet}}.
    \end{abstract}

}

\maketitle

% is not selected <OR> if conference mode is selected - because compsoc
% conference papers position the abstract like regular (non-compsoc)
\IEEEdisplaynontitleabstractindextext

% \fi
\IEEEpeerreviewmaketitle

\ifCLASSOPTIONcompsoc
    \IEEEraisesectionheading{\section{Introduction}\label{sec:introduction}}
\else
    \section{Introduction}
    \label{sec:introduction}
\fi

\IEEEPARstart{M}{ultivariate} time series forecasting (MTSF) plays an important role in various domains, including meteorology~\cite{price2025probabilistic}, traffic~\cite{zhang2011data}, economics~\cite{zhou2022robust} and so on.
By modeling complex temporal dependencies and inter-variate correlations, advanced MTSF methods~\cite{zhang2024self, liu2024itransformer,li2025mc} have demonstrated significant capabilities in predicting critical metrics and anticipating future trends.
Despite these significant advances, the practical deployment of MTSF models encounters a critical limitation: most existing approaches necessitate complete time series data, with observations available for all variates at each timestep.
This assumption may not hold in real-world scenarios, as time series data collected in practical applications are often inherently incomplete, frequently exhibiting missing observations due to sensor failures, communication interruptions, data corruption, or limited resources~\cite{wang2024deep,chen2021bayesian,xu2023conformal,zhang2023deta}.

\begin{figure}[tbp]
    \setlength{\abovecaptionskip}{5pt}  % 标题上方间距
    \setlength{\belowcaptionskip}{0.3pt}  % 标题下方间距
    \centerline{\includegraphics[width=0.96\columnwidth]{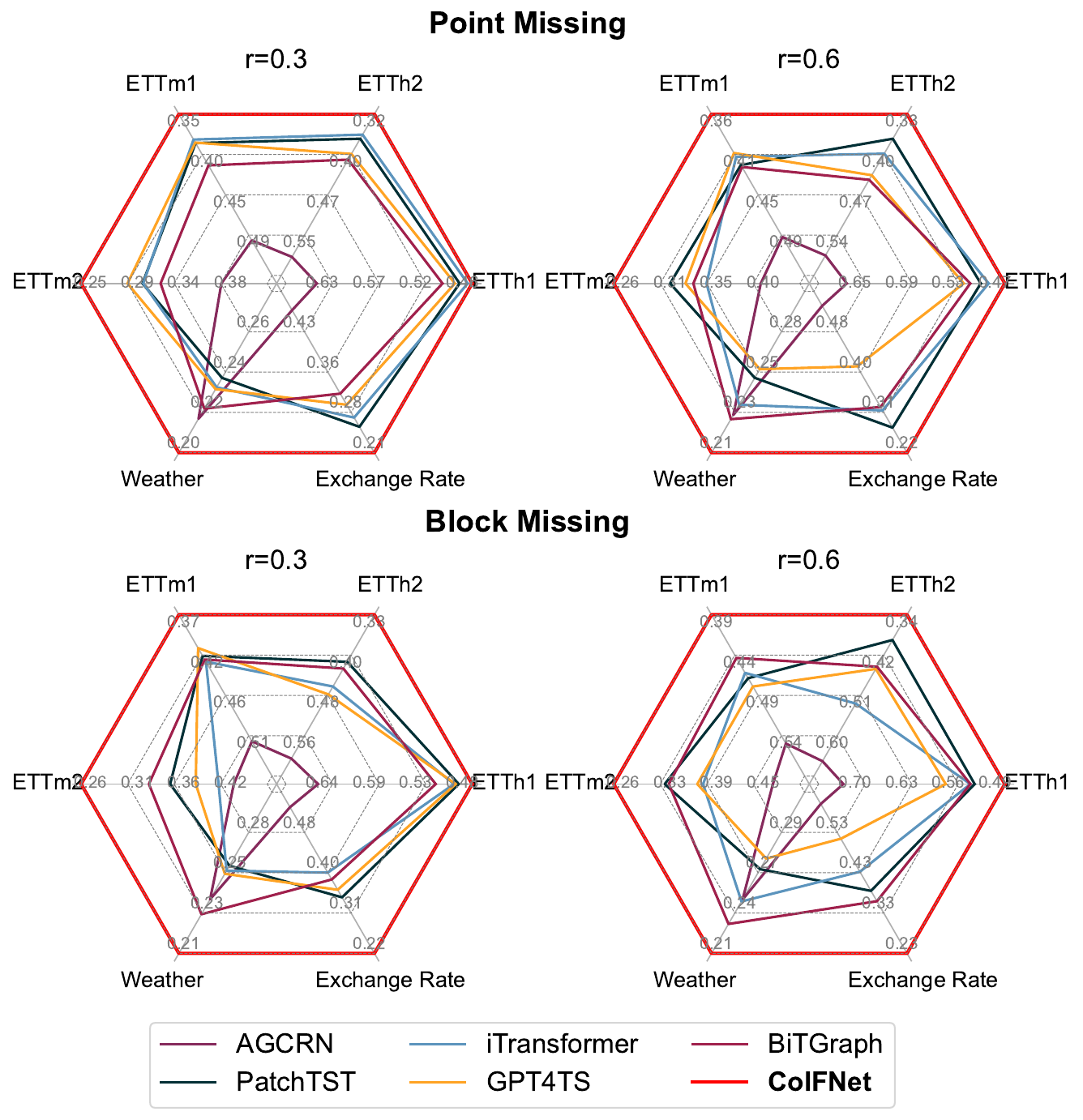}}
    % \centerline{\includegraphics[width=0.93\columnwidth]{fig/radar.pdf}}
    \caption{Performance comparison between our CoIFNet approach and state-of-the-art multivariate time series forecasting (MTSF) methods on six real-world datasets under different missing-data scenarios.}
    \label{fig:rad}
\end{figure}

Previous methods~\cite{li2017diffusion, yu2018spatio} generally adopt an impute-then-forecast pipeline, treating imputation and forecasting as distinct, sequential steps.
This two-stage framework leads to suboptimal predictions due to error propagation and misaligned optimization objectives.
To address this limitation, various one-stage approaches have been proposed.
One line of research formulates MTSF with missing data as an imputation task, where the forecasting window is treated as a block of missing values that need to be imputed~\cite{tashiro2021csdi, cao2018brits, luo2018multivariate}.
In contrast, the recently proposed BiTGraph approach~\cite{chen2024biased} performs forecasting directly on incomplete time series by leveraging mask-informed adaptive graphs to address missing values.
While these methods derive fully observed forecasting values through a single imputation or forecasting step, they overlook the complementary roles of imputation and forecasting, failing to leverage their mutual benefits for enhanced performance.
This highlights a critical open question in MTSF:
\begin{mdframed}
    [skipabove=4pt, innertopmargin=4pt, innerbottommargin=4pt]
    \textit{Can imputation and forecasting be unified within a single framework to achieve robust time series forecasting in the presence of missing data? }
\end{mdframed}
\vspace{-1pt}

In this work, we answer the question by proposing the \textbf{Co}llaborative \textbf{I}mputation-\textbf{F}orecasting \textbf{Net}work (\textbf{CoIFNet}).
\textbf{Technically}, CoIFNet first develops a Reversible Observed-value Normalization ({RevON}) module to to mitigate data distribution shifts across time steps within the input lookback window.
Then, the normalized observed values along with the corresponding mask matrix and timestamp embeddings are processed by sequential Cross-Timestep
Fusion ({CTF}) and Cross-Variate Fusion ({CVF}) modules to capture temporal dependencies robust to missing values.
At the end, the output series representations of the two sequential modules are projected to the original data space, where an imputation loss $\mathcal{L}_{\mathrm{I}}$ and a forecasting loss  $\mathcal{L}_{\mathrm{F}}$ are built for joint training.
\textbf{Theoretically}, we provide formal justifications grounded in the concept of {mutual information}~\cite{shannon1948mathematical}, proving how the CoIFNet learning objective enhances the performance bound for MTSF with missing data.
\textbf{Empirically}, we conduct extensive experiments on six real-world datasets across diverse missing-data scenarios (i.e., point missing and block missing), demonstrating CoIFNet's superior performance and computational efficiency compared with state-of-the-arts, as presented in  Fig.~\ref{fig:rad} and Fig.~\ref{fig:params}.

{To summarize, the main contributions of this work are threefold:}
\begin{itemize}
    \item {We propose CoIFNet, which unifies imputation and forecasting into a single framework to achieve robust
          MTSF in the presence of missing values.}
    \item We theoretically demonstrate the superiority of our one-stage CoIFNet framework over traditional two-stage approaches for handling missing values in time series forecasting.
    \item Extensive experiments on six challenging real-world datasets demonstrate the effectiveness and computational efficiency of our proposed approach across diverse missing-data scenarios.
\end{itemize}

The remainder of this paper is organized as follows. Section 2 reviews related work. Section 3 introduces our proposed CoIFNet method. Section 4 provides a theoretical analysis of CoIFNet, while Section 5 presents and discusses the experimental results. Section 6 concludes the paper. Additional experimental results are included in the Appendix.

\section{Related Work}

This section reviews existing multivariate time series forecasting (MTSF) methods designed for complete data, as well as representative approaches for handling missing values.

\subsection{Multivariate Time Series Forecasting}
MTSF has evolved from traditional statistical methods like ARIMA~\cite{nelson1998time, Bashir2018HandlingMD} to sophisticated deep learning architectures~\cite{Oreshkin2020NBEATS, wu2019graphwavenet}. Graph-based methods have emerged as particularly effective approaches for multivariate scenarios. AGCRN~\cite{bai2020adaptive} and MTGNN~\cite{wu2020connecting} construct adaptive graphs to model complex variate dependencies, demonstrating the importance of capturing both temporal dynamics and inter-variate relationships.
Recent research has revealed counterintuitive insights about architectural complexity. DLinear~\cite{zeng2023dlinear} shows that linear models with proper decomposition can outperform sophisticated methods, challenging assumptions about complexity and performance. Additionally, STID~\cite{shao2022spatial} demonstrates that auxiliary timestamp features significantly enhance forecasting accuracy by capturing seasonal dynamics.
Transformer-based models represent a significant breakthrough in long-sequence forecasting through self-attention mechanisms~\cite{zhou2021informer, wu2021autoformer}. PatchTST~\cite{nie2023patch} introduces patching techniques that improve efficiency and performance, while iTransformer~\cite{liu2024itransformer} employs inverted attention to treat variates as tokens rather than timesteps. The recent emergence of Large Language Models has introduced novel paradigms, with GPT4TS~\cite{zhou2023one} exploring how pre-trained language models can be adapted for time series analysis through prompt-based forecasting approaches~\cite{ma2024survey, jin2023time}.
Despite these advances, existing methods assume complete observations at each timestep. When facing missing values, most approaches resort to simple preprocessing strategies like mean imputation, creating suboptimal two-stage pipelines that fail to jointly optimize imputation and forecasting objectives.

\begin{figure}[tbp]
    \setlength{\abovecaptionskip}{4pt}  % 标题上方间距
    \centerline{\includegraphics[width=0.95\columnwidth]{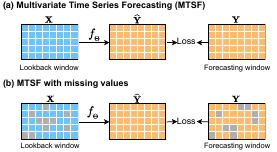}}
    \caption{
        % Comparison of evaluation strategies in time series forecasting with missing values. While BiTGraph only evaluates predictions on observed values in the test set, our method considers the accuracy of all forecasted points, preventing evaluation bias from increased missing rates. Blue and orange rectangles represent the lookback and forecasting windows respectively, with gray regions indicating missing values.
        Illustrations of Multivariate Time Series Forecasting (MTSF) with complete data \textbf{(a)} and with missing data \textbf{(b)}. The blue block and orange block represent the \textit{lookback window} and \textit{forecasting window}, respectively, with gray regions within each block indicating missing values.
    }
    \label{fig:problem}
\end{figure}

% \begin{figure}[!htbp]
%     \centerline{\includegraphics[width=0.9\columnwidth]{fig/Problem-def.pdf}}
%     \caption{Illustration of the PTSF problem. (a) Challenges in partial observations: complex missing patterns and disrupted temporal dependencies in real-world time series data. (b) Comparison between MTSF and PTSF: while MTSF learns from complete historical data, PTSF must handle partial observations while maintaining prediction accuracy.}
%     \label{fig:problem}
% \end{figure}

\begin{figure*}[t]
    \setlength{\abovecaptionskip}{-4pt}  % 标题上方间距
    \setlength{\belowcaptionskip}{0pt}  % 标题下方间距
    \centerline{\includegraphics[width=0.86\textwidth]{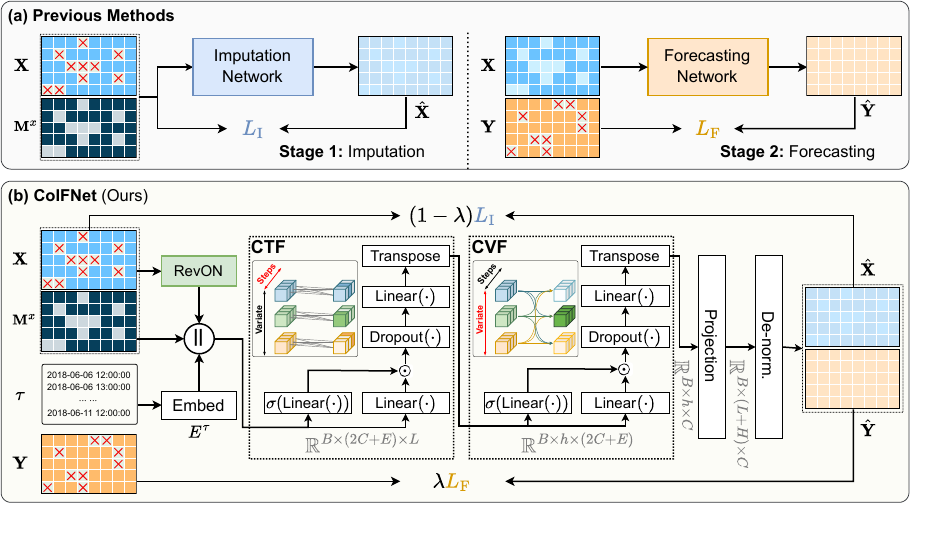}}
    % \caption{Framework overview of our proposed CoIFNet model. (a) Existing paradigm of ``Impute-then-Forecast". (b) Overall architecture of CoIFNet, .}
    \caption{\textbf{Illustration of CoIFNet.}
    Unlike previous methods that employ an input-then-forecast paradigm, CoIFNet formulates imputation and forecasting in a unified framework to achieve robust MTSF with missing values.
    {First}, CoIFNet introduces a Reversible Observed-value Normalization (RevON) module to mitigate data distribution shifts across time steps within the input lookback window $\mathbf{X}$. {Next}, the normalized values together with mask matrix $\mathbf{M}^x$ and temporal embeddings $E^{\tau}$ are fed into the sequential Cross-Timestep Fusion (CTF) and Cross-Variate Fusion (CVF) modules to learn temporal dependencies along the time step and variate dimensions. {Finally}, the output series representations are projected to the original data space, where an imputation loss $\mathcal{L}_{\mathrm{I}}$ and a forecasting loss  $\mathcal{L}_{\mathrm{F}}$ are developed based on the de-normalized representations $\hat{\mathbf{X}}$ and $\hat{\mathbf{Y}}$ as well as the input lookback window $\mathbf{X}$ and forecasting window (ground truth) $\mathbf{Y}$.
    }
    % first introduces an integrated framework that fuses mask indicators and temporal embeddings with rescaled historical sequences through RevON, emplopy a Cross-step(CSF) and Cross-Variable Fusion(CVF) module that learn the intra and inter temporal dependencies separately, leverages a Dual-Mixer Module to capture both intra-variable and inter-variable dependencies, and employs a joint projection layer for simultaneous imputation and forecasting optimization.}
    \label{fig:framework}
\end{figure*}

\subsection{MTSF with Missing Values}

Traditional approaches follow a two-stage paradigm that separates imputation and forecasting into sequential tasks. Statistical methods perform simple interpolation before applying forecasting models. Learning-based imputation methods like BRITS~\cite{cao2018brits}, PriSTI~\cite{liu2023pristi}, and SAITS~\cite{du2023saits} use RNNs or attention mechanisms to recover missing values by modeling temporal dependencies. However, this decoupled framework suffers from error propagation and misaligned optimization objectives between imputation and forecasting stages.
Recent efforts have explored end-to-end solutions to address these limitations. GRU-D~\cite{che2018recurrent} incorporates decay mechanisms for missing values but was originally designed for classification and lacks explicit forecasting optimization. Neural ODE approaches~\cite{rubanova2019latent, schirmer2022modeling} naturally handle irregular sampling but suffer from computational complexity and limited exploitation of temporal patterns.
More recent methods attempt unified treatment of missing values and forecasting. CSDI~\cite{tashiro2021csdi} reformulates forecasting as an imputation task using diffusion models, but exhibits discrepancy between training and inference stages.  Wang et al.~\cite{wang2024taskoriented} demonstrate that aligning imputation with forecasting objectives can significantly improve performance.
Although existing imputation-based approaches~\cite{tashiro2021csdi, cao2018brits, luo2018multivariate} can reconstruct complete inputs for forecasting, they fail to harness the full predictive potential of historical supervision signals.
BiTGraph~\cite{chen2024biased} proposes missing-aware adaptive graph construction but underutilizes temporal information and incurs high computational costs. S4M~\cite{peng2025s4m} integrates missing data handling into the structured state space sequence (S4) model architecture, effectively capturing the underlying temporal and
multivariate dependencies.
Although many one-stage forecasting models incorporate additional components to better utilize observed values, they often overlook the complementary roles of imputation and forecasting, thereby failing to leverage their mutual benefits for enhanced performance.
In addition, they typically require separate training of models for different miss rates, which limits their robustness \cite{11002729,liang2023knowledge,deng2024disentangling}.

\section{Methodology}
In this section, we introduce the Collaborative Imputation-Forecasting Network (CoIFNet), a one-stage MTSF framework designed to overcome the challenge of missing data.
Table~\ref{tab:notations} summarizes the notations used in this paper.

\subsection{Preliminaries}

\begin{table}[!t]
    \centering
    \caption{Key notations used in this paper}
    \label{tab:notations}
    \begin{tabular}{c|l}
        \toprule
        Notions        & \quad \quad \quad \quad \quad \quad Description                                  \\
        \hline
        $f_\Theta$                                & Neural network with parameters $\Theta$    \\
        $\mathcal{X} \in \mathbb{R}^{T \times D}$ & Time series with $T$ time steps and $D$ variates   \\
        $\mathcal{M} \in \{0,1\}^{T \times D}$    & Mask matrix for the the time series   \\
        % $T$                                       & Length of the time series                  \\
        % $L$                                       & Length of         \\
        % $H$                                       & Length of a forecasting window         \\
        % $D$                                       & Number of variates                           \\
        $\mathbf{X} \in \mathbb{R}^{L \times D}$  & A lookback window of length $L$    \\         
        $x_t \in \mathbb{R}^D$                    & Observation at time step $t$   \\
        $\mathbf{M}^x \in \{0,1\}^{L \times D}$   & Mask matrix for a lookback window \\
        $E^\tau \in \mathbb{R}^{L \times C}$      & Timestamp embed. for a lookback window  \\
        $\mathbf{Y} \in \mathbb{R}^{H \times D}$  & A forecasting window of length $H$       \\
        $\mathbf{M}^y \in \{0,1\}^{H \times D}$   & Mask matrix for a forecasting window  \\
        % $C$                                       & Dimension of timestamp embeddings             \\
        % $\tau \in \mathbb{R}^{L \times 2}$        & Timestamp features   \\
        \bottomrule
    \end{tabular}
\end{table}

Let $\mathcal{X} = \{\textbf{x}_{1}, \cdots, \textbf{x}_{T}\} \in \mathbb{R}^{T \times D}$ denote a multivariate time series  with $T$ time steps and $D$ variates.
As shown in Fig.~\ref{fig:problem}, each observation $\textbf{x}_{t} \in \mathbb{R}^{D}$ at time step $t$ may contain missing values due to sensor malfunctions, communication failures and so on. We represent these missing patterns of the multivariate time series using a mask matrix $\mathcal{M} \in \{0,1\}^{T \times D}$:
\begin{equation}
    \textbf{m}_{t,d} =
    \begin{cases}
        1, & \text{if } \textbf{x}_{t,d} \text{ is observed} \\
        0, & \text{otherwise}.
    \end{cases}
\end{equation}
where $\textbf{m}_{t,d}$ indicates the $t$-row, $n$-column element of $\mathcal{M}$.
It has been demonstrated that incorporating the mask matrix to the training process is of great importance, as it allows models to distinguish between actual zeros and missing values~\cite{du2023saits,cao2018brits}.
Given a lookback window (a.k.a. historical observations) $\mathbf{X}=\mathcal{X}_{t-L+1:t} \in \mathbb{R}^{L \times D}$ of length $L$   as well as its corresponding mask matrix $\mathbf{M}^x = \mathcal{M}_{t-L+1:t} \in \{0,1\}^{L \times D}$,
the goal of MTSF with missing values is to predict a forecasting window $\mathbf{Y}=\mathcal{X}_{t+1:t+H} \in \mathbb{R}^{H \times D}$ of length $H$, using a  neural network $f_\Theta: (\mathbf{X}, \mathbf{M}^x) \mapsto \mathbf{Y}$, where $\Theta$ denotes the learnable parameters.

\subsection{Overview}
Unlike most previous approaches that adopt a sequential input-then-forecast paradigm, CoIFNet unifies imputation and forecasting in a single framework for achieving robust MTSF in the presence of missing values.

An overview of our proposed CoIFNet is illustrated in Fig.~\ref{fig:framework}. Specifically, we first devise a \textit{Reversal Observed-value Normalization (RevON)} module to reduce data distribution shifts across time steps  within the input lookback window $\mathbf{X}$, yielding the normalized input $\bar{\mathbf{X}} \in \mathbb{R}^{L \times D}$.
% addressing the statistical bias typically introduced by missing values. 
Second, we concatenate $\bar{\mathbf{X}} \in \mathbb{R}^{L \times D}$ with the mask matrix $\mathbf{M}^x$ and timestamp embeddings $E^\tau \in \mathbb{R}^{L \times C}$ to produce $Z_{\text{in}} \in \mathbb{R}^{L \times (2D+C)}$. This composite input is processed by the \textit{Cross-Timestep Fusion (CTF)} module to obtain timestep-attentive representations $Z_{\text{CTF}} \in \mathbb{R}^{H \times (2D+C)}$.
Third, the \textit{Cross-Variate Fusion (CVF)} module takes $Z_{\text{CTF}}$ as input to obtain variate-attentive representations $Z_{\text{CVF}} \in \mathbb{R}^{H \times D}$.
Fourth, $Z_{\text{CVF}}$ are de-normalized in the original data space to obtain the imputed/reconstructed lookback window $\tilde{\mathbf{X}} \in \mathbb{R}^{ L \times D}$ and the predicted forecasting window $\tilde{\mathbf{Y}} \in \mathbb{R}^{ H \times D}$.
% both of which are de-normalized to yield the final outputs $\hat{\mathbf{X}}$ and $\hat{\mathbf{Y}}$.
Finally, an imputation loss $\mathcal{L}_{\mathrm{I}}$ together with a forecasting loss  $\mathcal{L}_{\mathrm{F}}$ are respectively built on ($\hat{\mathbf{X}}$, ${\mathbf{X}}$) and ($\hat{\mathbf{Y}}$, ${\mathbf{Y}}$) to jointly update the network parameters.

\subsection{Reversible Observed-value Normalization (RevON)}\label{sec:masknorm}
Real-world time series often exhibit non-stationarity, characterized by changes in statistical properties (such as mean and variance) across time steps. This phenomenon can lead to discrepancies across different time periods and affect a model's ability to generalize from past data to future events. Instance normalization has proven critically important for mitigating this issue. However, existing methods such as Reversible Instance Normalization (RevIN)~\cite{kim2021reversible}, which use affine transformations to mitigate discrepancies across historical sequences, rely on complete time series to compute statistics such as mean and variance.

To tackle this limitation, we devise Reversible Observed-value Normalization (RevON), which calculates normalization statistics using (partially) observed time series values. Given the mask matrix $\mathbf{M}^{x} = \{\textbf{m}_{1}, \cdots, \textbf{m}_{L}\}$ of a lookback window $\mathbf{X} = \{\textbf{x}_{1}, \cdots, \textbf{x}_{L}\}$,  RevON takes the form of
\begin{equation}
    \begin{aligned}
        \mathbb{E}[\mathbf{X}] & = \frac{1}{\sum_{t=1}^{L}\mathbf{M}^{x}_{t}}\sum_{t=1}^{L}\mathbf{M}^{x}_{t}\mathbf{x}_{t},                                \\
        \text{Var}[\mathbf{X}] & = \frac{1}{\sum_{t=1}^{L}\mathbf{M}^{x}_{t}}\sum_{t=1}^{L}\mathbf{M}^{x}_{t}(\mathbf{M}^{x}_{t}-\mathbb{E}[\mathbf{X}])^2. \\
    \end{aligned}\label{eq:norm}
\end{equation}
Thus, the normalized representations can be expressed as
\begin{equation}
    \bar{\mathbf{X}}=\left( \gamma \odot \left(\frac{\mathbf{X}-\mathbb{E}[\mathbf{X}]}{\sqrt{\text{Var}[\mathbf{X}]+\epsilon}}\right) + \beta \right)\odot \mathbf{M}^x,
\end{equation}
where $\gamma$ and $\beta$ are learnable weights and $\epsilon$ is a hyperparameter controlling stability.

\subsection{Cross-Timestep Fusion (CTF)}\label{sec:CTF}
Temporal context is critically important for processing incomplete time series, as it can compensate for the information distortion caused by missing values. Therefore, the Cross-Timestep Fusion (CTF) module of CoIFNet is first designed to fuse information along the timestep dimension to capture temporal dependencies within each variate of the multivariate time series.

Prior works have demonstrated the importance of leveraging timestamp information in time series to capture cyclical patterns~\cite{shao2022spatial, wang2024rethinking}. Motivated by this, we first extract the timestamp embeddings $E^\tau=\texttt{Embedding}(\tau)$ using the methodology introduced \cite{shao2022spatial}, and then we concatenate $E^\tau$ with the normalized time series data $\bar{\mathbf{X}}$ and the corresponding mask matrix $\mathbf{M}^x$ to obtain the input temporal representations:
\begin{equation}
    Z_{\mathrm{in}}=[\bar{\mathbf{X}} \parallel \mathbf{M}^x \parallel  E^\tau]  \in \mathbb{R}^{L \times (2D+C)},
\end{equation}
where
\begin{equation}
    E^\tau = [E_{\text{day}} \parallel  E_{\text{hour}}] \in \mathbb{R}^{L \times C},
\end{equation}
$E_{\text{day}} \in \mathbb{R}^{L \times C_d}$ and $E_{\text{hour}} \in \mathbb{R}^{L \times C_h}$ denote day-of-week and hour-of-day timestamp embeddings respectively.

Given this composite input representation $Z_{\text{in}}$, the CTF module learns temporal dependencies along the time step dimension through an adaptive gating mechanism:

\begin{equation}
    \begin{aligned}
        Z_{d}          & = \sigma(W_{\alpha}Z_{\text{in}} + b_{\alpha}) \odot (W_{d}Z_{\text{in}} + b_{d}), \\
        Z_{\text{CTF}} & = W_{p}\cdot\text{Dropout}\left(Z_{d}\right) + b_{p},
    \end{aligned}
\end{equation}
where $W_{\alpha}, W_d \in \mathbb{R}^{L \times h}$, $W_p \in \mathbb{R}^{h \times h}$ and $b_{\alpha}, b_d, b_p \in \mathbb{R}^{h}$ are learnable parameters, with $h$ representing the hidden layer dimension. The function $\sigma(\cdot)$ denotes the sigmoid activation and $\odot$ represents element-wise multiplication.
The sigmoid gate $\sigma(W_{\alpha}Z_{\text{in}} + b_{\alpha})$ produces values between 0 and 1, controlling how much information from each temporal feature is preserved.
Intuitively, this selective attention mechanism enables the model to focus on learning patterns relevant to observed values while reducing the influence of uncertain or noisy features that may result from missing data.

\subsection{Cross-Variate Fusion (CVF)}\label{sec:CVF}
Subsequently, the CVF module models the temporal dependencies along the variate dimension of $Z_{\text{CTF}} \in \mathbb{R}^{ h \times (2D+C)}$. Formally, CVF can be expressed as
\begin{equation}
    \begin{aligned}
        Z_{d}'         & = \sigma(W_{\alpha}'Z_{\text{CTF}} + b_{\alpha}') \odot (W_{d}'Z_{\text{CTF}} + b_{d}'), \\
        Z_{\text{CVF}} & = W_{p}'\cdot\text{Dropout}\left(Z_{d}'\right) + b_{p}',
    \end{aligned}
\end{equation}
where $W_{\alpha}', W_{d}' \in \mathbb{R}^{(2D+C) \times D}$, $W_{p}' \in \mathbb{R}^{D \times D}$, and $b_{\alpha}', b_{d}', b_{p}' \in \mathbb{R}^{D}$ are learnable parameters.
Similar to the CTF module, CVF employs a sigmoid-gated mechanism to adaptively focus on relevant cross-variate interactions while suppressing less informative ones. This adaptive approach is particularly beneficial when dealing with partially observed multivariate time series, as it helps mitigate the impact of missing values on cross-variate relationships.

Next, we project $Z_{\text{CVF}}$ to the original (low-dimensional) data space to simultaneously obtain the reconstructed historical values $\tilde{\mathbf{X}}$ and the predicted future values $\tilde{\mathbf{Y}}$:
\begin{equation}
    [\tilde{\mathbf{X}}, \tilde{\mathbf{Y}}] = W_o Z_{\text{CVF}} + b_o,\label{eq:projection}
\end{equation}
where $W_o \in \mathbb{R}^{h \times (L+H)}$ and $b_o \in \mathbb{R}^{L+H}$ are learnable parameters, with $L$ and $H$ representing the historical and forecast horizon lengths respectively.
% The outputs $\tilde{\mathbf{X}} \in \mathbb{R}^{ L \times D}$ and $\tilde{\mathbf{Y}} \in \mathbb{R}^{ H \times D}$ represent the reconstructed historical series and forecasted future series in the normalized space.
% However, the input data have different statistics than the original distribution, and by observing only the normalized input ˆx(i), it is difficult to capture the original distribution of the input x(i). Thus,
% to make this easier for the model, we explicitly return the non-stationary properties removed from
% the input data to the model output by reversing the normalization step at a symmetric position, the
% output layer.
% However, the normalized output lacks the statistical properties of the original data distribution. This discrepancy poses a challenge since the model must learn to generate predictions in a normalized space while the evaluation occurs in the original data space. 
% However, the normalized output $[\tilde{\mathbf{X}}, \tilde{\mathbf{Y}}]$ lacks the statistical information of the original data distribution $\mathbf{X}$. 
Finally, we apply a de-normalization process of RevON to align the statistical information of $[\tilde{\mathbf{X}}, \tilde{\mathbf{Y}}]$ to the original data distribution:

\begin{equation}
    [\hat{\mathbf{X}}, \hat{\mathbf{Y}}]  = \sqrt{\text{Var}[\mathbf{X}]+\epsilon}\odot \left( \frac{[\tilde{\mathbf{X}} , \tilde{\mathbf{Y}}]- \beta}{\gamma} \right)+\mathbb{E}[\mathbf{X}],\label{eq:denorm}
\end{equation}
where $\hat{\mathbf{X}}=\{\hat{\textbf{x}}_1,...,\hat{\textbf{x}}_L\}$, $\hat{\mathbf{Y}}=\{\hat{\textbf{y}}_1,...,\hat{\textbf{y}}_H\}$ and $\gamma$ and $\beta$ are the learnable parameters in Eq.~\ref{eq:norm}.
% Our ablation study results confirm that this step plays a crucial role in mitigating the data drift issue and yielding more reliable prediction results.

\begin{algorithm}[t]
    \caption{Training pipeline of CoIFNet}
    \label{alg:fr_learning}
    \begin{algorithmic}[1]
        \item[\textbf{Input:}]  Multivariate time series $\mathcal{X}$, mask matrix $\mathcal{M}$, timestamp features $\mathcal{T}$, hyperparameter $\lambda$
        \item[\textbf{Output:}] Trained model parameters $\Theta$
        \STATE Initialize model parameters $\Theta$
        \WHILE{not converged}
        \STATE Sample $(\mathbf{X}, \mathbf{Y}) \sim \mathcal{X}$, $(\mathbf{M}^x, \mathbf{M}^y) \sim \mathcal{M}$, $\tau \sim \mathcal{T}$
        \STATE $\bar{\mathbf{X}} \leftarrow \text{RevON}(\mathbf{X}, \mathbf{M}^x)$
        \STATE $E^\tau \leftarrow \texttt{Embedding}(\tau)$
        \STATE $Z_\text{in} \leftarrow [\tilde{\mathbf{X}} \| \mathbf{M}^x \| E^\tau]$
        \STATE $Z_\text{CTF} \leftarrow \text{CTF}(Z_\text{in})$
        \STATE $Z_\text{CVF} \leftarrow \text{CVF}(Z_\text{CTF})$
        \STATE $[\bar{\mathbf{X}}, \tilde{\mathbf{Y}}] \leftarrow \texttt{Projection}(Z_\text{CVF})$
        \STATE $[\hat{\mathbf{X}}, \hat{\mathbf{Y}}] \leftarrow \text{De-Norm.}([\tilde{\mathbf{X}}, \tilde{\mathbf{Y}}])$
        \STATE Calculate  $\mathcal{L}_\text{I}$ and $\mathcal{L}_\text{F}$
        \STATE Update $\Theta$ by minimizing $\mathcal{L} = (1-\lambda)\mathcal{L}_\text{I} + \lambda\mathcal{L}_\text{F}$
        \ENDWHILE
        \RETURN $\Theta$
    \end{algorithmic}
\end{algorithm}

\subsection{Learning Objective}
% To effectively train our CoIFNet, we employ a combined loss function that simultaneously addresses both the reconstruction of missing historical values and the forecasting of future values. This unified approach allows the model to leverage patterns discovered during imputation to enhance forecasting accuracy, while also ensuring that the imputation task benefits from the model's understanding of future trends.
The overall learning objective of the proposed CoIFNet is defined as a weighted combination of the imputation loss $\mathcal{L}_{\mathrm{I}} $ and the prediction loss $\mathcal{L}_{\mathrm{F}}$:
\begin{equation}
    \mathcal{L} = (1 - \lambda) \mathcal{L}_{\mathrm{I}} + \lambda \mathcal{L}_{\mathrm{F}},
    \label{eq.loss}
\end{equation}
where $\lambda \in [0, 1]$ is a hyperparameter controling the relative importance of the two losses:
% The reconstruction loss $\mathcal{L}_{\text{imp}}$ and prediction loss $\mathcal{L}_{\text{fore}}$ are defined as follows:
\begin{equation}
    \begin{aligned}
        \mathcal{L}_{\mathrm{I}} & =\frac{\sum_{t}\mathbf{M}^x_{t}\cdot|\mathbf{X}_{t}-\hat{\mathbf{X}}_{t}|}{\sum_{t}\mathbf{M}^x_{t}}, \quad t=1,...,L  \\
        \mathcal{L}_{\mathrm{F}} & =\frac{\sum_{h}\mathbf{M}^y_{h}\cdot|\mathbf{Y}_{h}-\hat{\mathbf{Y}}_{h}|}{\sum_{h}\mathbf{M}^y_{h}}, \quad h=1,...,H.
    \end{aligned}
\end{equation}
% This formulation allows us to balance the focus between accurate historical reconstruction and precise future forecasting, depending on the specific requirements of the application.
% where $\textbf{y}_t$, $\hat{\mathbf{y}}_t$, $\textbf{m}_t^x$ and $\textbf{m}_t^y$ denote the $t$-th rows of $\textbf{X}$, $\hat{\mathbf{X}}$ and $[\tilde{\mathbf{X}}$.
% $\mathbf{M}^y = \mathcal{M}_{t+1:t+H} \in \{0,1\}^{H \times D}$ indicates the mask matrix for the forecasting horizon. 
% Note that we only consider observed values in the loss calculation. 
In this way, CoIFNet integrates imputation and forecasting into a unified framework, enabling end-to-end learning of temporal dependencies from incomplete time series data.
As a lightweight network architecture, CoIFNet ensures lower computational complexity and faster execution speeds (as  demonstrated in Section~\ref{exp:comput}), which makes it highly suitable for deployment in resource-constrained environments. The training pipeline for CoIFNet is detailed in \textbf{Algorithm}~\ref{alg:fr_learning}.

\section{Theoretical Analysis}
\label{sec:info_theory_analysis}

In this section, we provide formal justifications grounded in the concept of \textit{mutual information}~\cite{shannon1948mathematical}, demonstrating how the CoIFNet learning objective enhances the performance bound for MTSF in the presence of missing data.

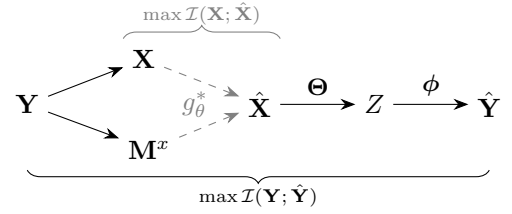
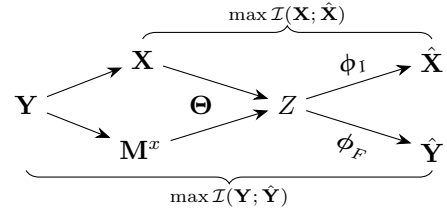
\begin{figure}[t]
    \centering
    \setlength{\abovecaptionskip}{0pt}  % 标题上方间距
    \setlength{\belowcaptionskip}{0pt}  % 标题下方间距
    % Define some styles
    \tikzstyle{var}=[circle, draw, minimum size=0.1cm, inner sep=1pt]
    \tikzstyle{arr}=[-{Stealth[length=2mm, width=1.5mm]}]
    \tikzstyle{dashedarr}=[arr, dashed]
    \tikzstyle{brace}=[decorate, decoration={brace, amplitude=5pt, raise=1pt}]
    \tikzstyle{annText}=[midway, above=3pt, font=\scriptsize]
    \tikzstyle{annTextBelow}=[midway, below=3pt, font=\scriptsize]

    \begin{subfigure}{0.95\columnwidth}
        \centering
        \setlength{\abovecaptionskip}{0pt}  % 标题上方间距
        \setlength{\belowcaptionskip}{0pt}  % 标题下方间距
        \begin{tikzpicture}[node distance=0.8cm and 1.2cm]
            % Nodes (same as a)
            \node (hatX) {$\hat{\mathbf{X}}$};

            \node (X) [left=1.0cm of hatX, yshift=0.6cm] {$\mathbf{X}$};
            \node (M) [left=0.8cm of hatX, yshift=-0.6cm] {$\mathbf{M}^x$};

            \node (dd) [left=0.3cm of hatX, gray] {$g_\theta^*$};

            \draw[arr, dashed, gray] (M) -- (hatX);
            \draw[arr, dashed, gray] (X) -- (hatX);

            \node (Z) [right=1cm of hatX] {$Z$};

            \node (Yhata) [right=1cm of Z] {$\hat{\mathbf{Y}}$};

            \node (Y) [left=1cm of X, yshift=-0.6cm] {$\mathbf{Y}$};
            \draw[arr] (Y) -- (X);
            \draw[arr] (Y) -- (M);

            % Arrows (same as a)
            \draw[arr] (hatX) -- node[above, midway, font=\small] {$\boldsymbol{\Theta}$} (Z);

            \draw[arr] (Z) -- node[above, sloped, midway, font=\small] {$\boldsymbol{\phi}$} (Yhata);

            \path (X.north) ++(0,0) coordinate (common_y_top_a);
            \coordinate (brace_a_L_top) at (X.west |- common_y_top_a);
            \coordinate (brace_a_R_top) at (hatX.east |- common_y_top_a);
            \draw[brace, gray] (brace_a_L_top) -- (brace_a_R_top) node[annText] {$\max \I(\mathbf{X}; \hat{\mathbf{X}})$};

            \path (M.south) ++(0,0) coordinate (common_y_top_a);
            \coordinate (brace_a_L_top) at (Y.center |- common_y_top_a);
            \coordinate (brace_a_R_top) at (Yhata.center |- common_y_top_a);
            \draw[brace, decoration={mirror}] (brace_a_L_top) -- (brace_a_R_top) node[annTextBelow] {$\max \I({\mathbf{Y}}; \hat{\mathbf{Y}})$};

        \end{tikzpicture}
        \caption{Traditional impute-then-forecast approaches}
        \label{fig:two}
    \end{subfigure}%
     \vspace{1ex}
    \begin{subfigure}{0.95\columnwidth}
        \centering
        \setlength{\abovecaptionskip}{0pt}  % 标题上方间距
        \setlength{\belowcaptionskip}{5pt}  % 标题下方间距
        \begin{tikzpicture}[node distance=0.8cm and 1.2cm]
            \node (Z) {$Z$};

            \node (X) [left=1.4cm of Z, yshift=0.6cm] {$\mathbf{X}$};
            \node (M) [left=1.3cm of Z, yshift=-0.6cm] {$\mathbf{M}^x$};

            \node (dd) [left=0.6cm of hatX] {$\boldsymbol{\Theta}$};
            \draw[arr] (M) -- (Z);
            \draw[arr] (X) -- (Z);

            \node (Xhat) [right=1.4cm of Z, yshift=0.6cm] {$\hat{\mathbf{X}}$};
            \node (Yhata) [right=1.4cm of Z, yshift=-0.6cm] {$\hat{\mathbf{Y}}$};

            \node (Y) [left=1cm of X, yshift=-0.6cm] {$\mathbf{Y}$};
            \draw[arr] (Y) -- (X);
            \draw[arr] (Y) -- (M);

            \draw[arr] (Z) -- node[above, sloped, midway, font=\small] {$\boldsymbol{\phi}_I$} (Xhat);
            \draw[arr] (Z) -- node[below, sloped, midway, font=\small] {$\boldsymbol{\phi}_F$} (Yhata);

            \path (X.north) ++(0,0) coordinate (common_y_top_a);
            \coordinate (brace_a_L_top) at (X.center |- common_y_top_a);
            \coordinate (brace_a_R_top) at (Xhat.center |- common_y_top_a);
            \draw[brace] (brace_a_L_top) -- (brace_a_R_top) node[annText] {$\max \I(\mathbf{X}; \hat{\mathbf{X}})$};

            \path (M.south) ++(0,0) coordinate (common_y_top_a);
            \coordinate (brace_a_L_top) at (Y.center |- common_y_top_a);
            \coordinate (brace_a_R_top) at (Yhata.center |- common_y_top_a);
            \draw[brace, decoration={mirror}] (brace_a_L_top) -- (brace_a_R_top) node[annTextBelow] {$\max \I({\mathbf{Y}}; \hat{\mathbf{Y}})$};

        \end{tikzpicture}
        \caption{Our proposed CoIFNet framework}
        \label{fig:coif}
    \end{subfigure}%
    % \caption{Comparison of information flow in MTSF with missing values. (a) Two-stage pipeline: missingness indicator $\mathbf{M}^x$ (dashed gray) become inaccessible during forecasting, breaking the Markov chain continuity from $\mathbf{Y}$ to $\hat{\mathbf{Y}}$. (b) CoIFNet: joint learning preserves access to both $\mathbf{X}$ and $\mathbf{M}^x$, maintaining complete information flow for both tasks.}
    \caption{\textbf{Qualitative comparison of information flow between traditional impute-then-forecast approaches and our proposed CoIFNet framework}. \textbf{(a)} The mask matrix $\mathbf{M}^x$ used in the imputation stage becomes inaccessible during forecasting, breaking the Markov chain continuity from $\mathbf{Y}$ to $\hat{\mathbf{Y}}$. \textbf{(b)} CoIFNet unifies imputation and forecasting in a single framework, enabling Markov chain continuity and preserving all information critical for accurate time series forecasting in the presence of missing data.}
    \label{fig:mtl_comparison}
\end{figure}

\subsection{Information Loss in the Two-stage Pipeline}

Traditional approaches to missing data in time series follow a two-stage pipeline: imputing missing values then performing forecasting on the imputed data. As depicted in Fig.~\ref{fig:mtl_comparison}(a), the mask matrix $\mathbf{M}^x$ used in the imputation stage becomes inaccessible during forecasting, breaking the Markov chain continuity from $\mathbf{Y}$ to $\hat{\mathbf{Y}}$. According to the Data Processing Inequality (DPI)~\cite{cover2005} theory, we have:
\begin{equation}\label{eq:DPI}
    \I(\mathbf{Y}; \mathbf{X}, \mathbf{M}^x) \geq \I(\mathbf{Y}; \hat{\mathbf{X}}).
\end{equation}
where $\mathcal{I}(\cdot)$ denotes the the amount of mutual information.
The DPI Equation~(\ref{eq:DPI}) holds with equality \textit{only} if no information about $\mathbf{Y}$ is lost during processing.
However, the two-stage pipeline inherently causes information loss and results in a strict DPI inequality due to:
1) the imputation stage optimizes $\hat{\mathbf{X}}$ for reconstruction instead of forecasting, discarding a part of information critical for accurate forecasting~\cite{wang2024taskoriented}, and 2) the mask matrix $\mathbf{M}^x$, which holds valuable forecasting signals, is discarded after imputation~\cite{van2023mim}.

\subsection{Information Preservation via Joint Optimization}
Instead of following the restrictive path $(\mathbf{X}, \mathbf{M}^x) {\color{black}\xrightarrow[]{g_\theta^*}}\hat{\mathbf{X}} \xrightarrow[]{f_{\Theta, \phi}} \hat{\mathbf{Y}}$, we propose to learn from $(\mathbf{X}, \mathbf{M}^x)$ and produce both $\hat{\mathbf{X}}$ and $\hat{\mathbf{Y}}$ simultaneously. This preserves access to all input information throughout the learning process while maintaining the benefits of both imputation and forecasting objectives (as illustrated in Fig.~\ref{fig:mtl_comparison}(b)).
To formalize this intuition, we relate mutual information to the Mean Absolute Error (MAE) loss. The choice of MAE is motivated by its inherent robustness, which makes it particularly well-suited for time series forecasting with missing data, as highlighted by~\cite{cheng2024robusttsf}. This robustness stems from MAE's equivalence to maximizing likelihood under an assumption of Laplacian-distributed prediction errors, providing the theoretical foundation for our analysis. We employ a variational distribution $q(\mathbf{Y}|\hat{\mathbf{Y}})$ to approximate the true conditional probability $p(\mathbf{Y}|\hat{\mathbf{Y}})$, leading to the following proposition.

\begin{mdframed}[linecolor=white]
    \label{prop:info_preservation_revised}
    \textbf{\textit{Proposition 1.}}
    \textit{Under the assumption that prediction errors follow a Laplacian distribution, the mutual information $\I(\mathbf{Y}; \hat{\mathbf{Y}})$ between the ground truth values $\mathbf{Y}$ and predictions $\hat{\mathbf{Y}}$ is lower-bounded by:}
    \begin{equation*}
        \I(\mathbf{Y}; \hat{\mathbf{Y}}) \geq \mathcal{C} - \frac{1}{b}\E \left[\left\| \mathbf{Y} - \hat{\mathbf{Y}}\right\|_1\right],
    \end{equation*}
    \textit{where $\mathcal{C} = \Hc(\mathbf{Y}) - \log(2b)$, $b > 0$ is the scale parameter of the variational Laplacian distribution  $q(\mathbf{Y}|\hat{\mathbf{Y}})$, and $\Hc(\mathbf{Y})$ is the entropy of $\mathbf{Y}$.}
\end{mdframed}

\begin{IEEEproof}[\textbf{Proof}]
    Denote the mutual information of the ground truth $\mathbf{Y}$ and prediction $\hat{\mathbf{Y}}$ as:
    \begin{equation}
        \begin{aligned}
            \I(\mathbf{Y}; \hat{\mathbf{Y}}) = & \sum_{\mathbf{Y},\hat{\mathbf{Y}}} p(\mathbf{Y}, \hat{\mathbf{Y}}) \log \frac{p(\mathbf{Y}, \hat{\mathbf{Y}})}{p(\mathbf{Y}) p(\hat{\mathbf{Y}})} \\
            =                                  & \sum_{\mathbf{Y},\hat{\mathbf{Y}}} p(\mathbf{Y}, \hat{\mathbf{Y}}) \log \frac{p(\mathbf{Y}, \hat{\mathbf{Y}})}{p(\hat{\mathbf{Y}})}               \\
                                               & \quad - \sum_{\mathbf{Y},\hat{\mathbf{Y}}}  p(\mathbf{Y}, \hat{\mathbf{Y}})   \log p(\mathbf{Y})                                                  \\
            =                                  & \sum_{\mathbf{Y},\hat{\mathbf{Y}}} p(\hat{\mathbf{Y}})p(\mathbf{Y}| \hat{\mathbf{Y}}) \log p(\mathbf{Y}| \hat{\mathbf{Y}})                        \\
                                               & \quad - \sum_{\mathbf{Y},\hat{\mathbf{Y}}}  p(\mathbf{Y}, \hat{\mathbf{Y}})   \log p(\mathbf{Y})                                                  \\
            =                                  & \Hc(\mathbf{Y}) - \Hc(\mathbf{Y} | \hat{\mathbf{Y}})                                                                                              \\
            =                                  & \Hc(\mathbf{Y}) + \E_{\mathbf{Y},\hat{\mathbf{Y}}}[\log p(\mathbf{Y} | \hat{\mathbf{Y}})]
        \end{aligned}
    \end{equation}

    Since the true conditional distribution $p(\mathbf{Y} | \hat{\mathbf{Y}})$ is intractable, we introduce a variational distribution $q(\mathbf{Y} | \hat{\mathbf{Y}})$ to approximate it. The mutual information can then be expressed as:

    \begin{equation}
        \begin{aligned}
            \I(\mathbf{Y}; \hat{\mathbf{Y}})= & \Hc(\mathbf{Y}) + \E_{\hat{\mathbf{Y}}}[\log q(\mathbf{Y} | \hat{\mathbf{Y}})]                                                  \\
                                              & \quad +   \E_{\hat{\mathbf{Y}}}[\mathcal{D}_{\text{KL}}(p(\mathbf{Y} | \hat{\mathbf{Y}}) \| q(\mathbf{Y} | \hat{\mathbf{Y}})) ] \\
            \geq                              & \Hc(\mathbf{Y}) + \E_{\mathbf{Y},\hat{\mathbf{Y}}}[\log q(\mathbf{Y} | \hat{\mathbf{Y}})]
        \end{aligned}
    \end{equation}
    The inequality holds due to the non-negativity of the KL divergence $\mathcal{D}_{\text{KL}}(p \| q)$.
    We assume $q(\mathbf{Y} | \hat{\mathbf{Y}})$ is a Laplacian distribution centered at $\hat{\mathbf{Y}}$ with scale $b > 0$, i.e.,

    \begin{equation} \label{eq:laplace_pdf_app}
        q(\mathbf{Y} | \hat{\mathbf{Y}}) = \frac{1}{2b} \exp\left(-\frac{\|\mathbf{Y} - \hat{\mathbf{Y}}\|_1}{b}\right).
    \end{equation}
    Substituting the logarithm of Eq.~\eqref{eq:laplace_pdf_app} into the lower bound:

    \begin{equation}
        \begin{aligned}
            \I(\mathbf{Y}; \hat{\mathbf{Y}}) \geq & \Hc(\mathbf{Y}) + \E \left[\log q(\mathbf{Y} | \hat{\mathbf{Y}})\right]                                                      \\
            =                                     & \Hc(\mathbf{Y}) + \E \left[\log\left(\frac{1}{2b}\right) - \frac{\left\| \mathbf{Y} - \hat{\mathbf{Y}}\right\|_1}{b} \right] \\
            =                                     & \Hc(\mathbf{Y})  - \log(2b) - \frac{1}{b}\E \left[\left\| \mathbf{Y} - \hat{\mathbf{Y}}\right\|_1\right]                     \\
            =                                     & \mathcal{C} - \frac{1}{b}\E \left[\left\| \mathbf{Y} - \hat{\mathbf{Y}}\right\|_1\right]                                     \\
        \end{aligned}
    \end{equation}
    where $\mathcal{C} = \Hc(\mathbf{Y})  - \log(2b)$ is constant.
    Thus, minimizing MAE is tantamount to maximizing a variational lower bound on $\I(\mathbf{Y}; \hat{\mathbf{Y}})$ under these Laplacian assumptions.
\end{IEEEproof}

Proposition 1 reveals that minimizing the MAE directly maximizes the mutual information between the predictions and the ground truth, leading to the following corollary:

\begin{mdframed}[linecolor=white]
    \label{prop:info_preservation_revised}
    \textbf{\textit{Corollary 1.}}
    \textit{The mutual information for the imputation task satisfies: }
    \begin{equation*}
        \I(\mathbf{X}; \hat{\mathbf{X}}) \geq \mathcal{C}' - \frac{1}{b'}\E \left[\left\| \mathbf{X} - \hat{\mathbf{X}}\right\|_1\right],
    \end{equation*}
    \textit{where $\mathcal{C}'$ and $b' > 0$ are corresponding constants for $\mathbf{X}$}
\end{mdframed}

This indicates that joint optimization of imputation and forecasting preserves information for the both tasks.

\begin{mdframed}[linecolor=white]
    \label{prop:info_preservation_joint_revised}
    \textbf{\textit{Proposition 2.}}
    \textit{Let $Z$ be an internal representation learned from input $(\mathbf{X}, \mathbf{M}^x)$ through joint optimization. The total mutual information between $Z$ and all relevant variables is lower-bounded by:}
    \begin{equation*}
        \begin{aligned}
            \I(\mathbf{X}; Z) + & \I(\mathbf{Y}; Z) \\& \geq \mathcal{C}'' - \left(\frac{\E \left[\left\| \mathbf{Y} - \hat{\mathbf{Y}}\right\|_1\right]}{b} + \frac{\E \left[\left\| \mathbf{X} - \hat{\mathbf{X}}\right\|_1\right]}{b'}\right)
        \end{aligned}
    \end{equation*}
    \textit{where $\mathcal{C}'' = \mathcal{C} + \mathcal{C}'$ combines the entropy terms of the imputation and forecasting tasks.}
\end{mdframed}

\begin{IEEEproof}[\textbf{Proof}]
    We first bound the mutual information $\I(Z; \mathbf{X}, \mathbf{M}^x, \mathbf{Y})$ by decomposing it into components related to imputation and forecasting tasks.
    Since $Z$ is learned to optimize both tasks, we have:
    \begin{equation*}
        \I(\mathbf{X}, \mathbf{M}^x, \mathbf{Y}; Z) \geq \I( \mathbf{X}; Z) + \I(\mathbf{Y}; Z)
    \end{equation*}
    Thus,
    \begin{equation*}
        \I(\mathbf{X}; Z) + \I(\mathbf{Y}; Z) \geq \I(\mathbf{X}; \hat{\mathbf{X}}) + \I(\mathbf{Y}; \hat{\mathbf{Y}})
    \end{equation*}
    According to Proposition 1 and Corollary 1, we have
    \begin{equation*}
        \begin{aligned}
            \I(\hat{\mathbf{X}}; \mathbf{X}) + \I(\hat{\mathbf{Y}}; \mathbf{Y})
             & \geq \left(\mathcal{H}(\mathbf{X}) - \log(2b') - \frac{1}{b'}\mathbb{E}[|\mathbf{X} - \hat{\mathbf{X}}|]\right)                               \\
             & \quad + \left(\mathcal{H}(\mathbf{Y}) - \log(2b) - \frac{1}{b}\mathbb{E}[|\mathbf{Y} - \hat{\mathbf{Y}}|]\right)                              \\
             & = \mathcal{C}'' - \left(\frac{\mathbb{E}[|\mathbf{Y} - \hat{\mathbf{Y}}|]}{b} + \frac{\mathbb{E}[|\mathbf{X} - \hat{\mathbf{X}}|]}{b'}\right)
        \end{aligned}
    \end{equation*}
    where $\mathcal{C}'' = \mathcal{H}(\mathbf{Y}) + \mathcal{H}(\mathbf{X}) - \log(2b) - \log(2b')$ and $b, b' > 0$ are constants.
    So, minimizing imputation and forecasting MAE errors jointly means maximizes the lower bound on the mutual information captured by the representation $Z$.
\end{IEEEproof}

The representation $Z$ has direct access to both observed values and the mask matrix $\mathbf{M}^x$, ensuring no predictive information is discarded. Our CoIFNet framework implements this principle through the composite loss $\mathcal{L} = (1-\lambda) \mathcal{L}_{\mathrm{I}} + \lambda \mathcal{L}_{\mathrm{F}}$, where $\lambda \in [0,1]$ balances the imputation loss $\mathcal{L}_{\mathrm{I}}$ and forecasting loss $\mathcal{L}_{\mathrm{F}}$.
This ensures that the gradients from both objectives directly impact the shared representation  $Z$, thereby maximizing information preservation for robust and accurate time series forecasting.

\subsection{Discussion on the Underlying Assumption} It is worth noting that our formal connection between the MAE loss and mutual information in \textit{Proposition 1} is established under the assumption of a Laplacian distribution for the prediction errors. This choice is motivated by the direct correspondence between minimizing MAE and maximizing the variational lower bound on mutual information, a common approach in information-theoretic analyses of robust models.
However, the core principle motivating our joint optimization framework is more general. A parallel argument can be formulated for other common loss-distribution pairs. For instance, minimizing the Mean Squared Error (MSE) loss is tantamount to maximizing the mutual information lower bound under a Gaussian error assumption. Thus, while our specific derivation focuses on the empirically robust MAE/Laplacian case, the fundamental theoretical advantage of our unified CoIFNet framework—preserving predictive information by avoiding the information bottleneck inherent in two-stage pipelines (as illustrated in Fig. \ref{fig:mtl_comparison})—holds more broadly. The joint learning objective encourages the model to retain all features from both the observed data and the mask matrix that are pertinent to both tasks, a principle that is not confined to a single choice of error distribution.

\section{Experiments}

In this section, we conduct experiments on six real-world time series forecasting datasets under
different missing-data patterns to evaluate the effectiveness and efficiency of our CoIFNet framework.
We first compare the performance of CoIFNet with state-of-the-art approaches in handling diverse missing-data scenarios. Then, we perform ablation studies and case analysis to investigate the impacts of different model design decisions. The results under the point or block missing rate of 0.1 are provided at the end.

\subsection{Experiment Settings}

\noindent{\textbf{Datasets}:
% Following common benchmarks in multivariate time series forecasting,  each presenting unique characteristics and challenges
We include 6 most commonly used time series forecasting datasets in our experiments. Table~\ref{tb1:dataset} provides basic statistics for these datasets.

\begin{itemize}
    % \item \textbf{ECL:} Hourly electricity consumption data for 321 clients from 2011 to 2014. The data is split into training, validation, and test sets using a 12/2/2 month ratio.
    \item \textbf{Weather}\footnote{\url{https://www.bgc-jena.mpg.de/wetter/}}: Meteorological data consisting of 21 indicators recorded at 10-minute intervals from 07/2020 to 07/2021.
    \item \textbf{Exchange Rate}: Daily exchange rates for 8 countries from 1990 to 2016.
    \item \textbf{ETT (ETTh1, ETTh2, ETTm1, ETTm2)}\footnote{\url{https://github.com/zhouhaoyi/ETDataset}}: Electricity transformer temperatures and power load features at 1-hour (ETTh1, ETTh2) and 15-minute (ETTm1, ETTm2) intervals.

          % \item ETT (Electricity Transformer Temperature)\footnote{\url{https://github.com/zhouhaoyi/ETDataset}}: This dataset contains the oil temperature and six power load features from electricity transformers from July 2016 to July 2018. It can be further divided into four sub-datasets (ETTh1, ETTh2, ETTm1, and ETTm2) based on data collection granularity and regions.
          % \item Electricity\footnote{\url{https://archive.ics.uci.edu/ml/datasets/ElectricityLoadDiagrams20112014}}: This dataset collects the electricity consumption of 321 clients from 2012 to 2014.
          % \item Weather\footnote{\url{https://www.bgc-jena.mpg.de/wetter/}}: This dataset records various climate indicators such as temperature, humidity, and wind speed for 2020 in Germany.
\end{itemize}

% These datasets provide a diverse range of applications for evaluating the proposed method in multivariate time series forecasting with missing values across various domains. 

\noindent{\textbf{Missing Patterns}:
We consider two missing patterns in our experiments.
% The models are then evaluated under two types of missing values scenarios, each with two different missing rates $r=\{0.3, 0.6\}$:
% to the complete time series $\mathcal{X}^c$ before windowing

\begin{itemize}
    \item \textbf{Point Missing}: A certain rate of points in time series are randomly masked as missing values.
          % , simulating random sensor failures or data corruption.

    \item \textbf{Block Missing}: A certain rate of blocks $B_{t,c} \in \mathbb{R}^{l_t \times l_c}$  in time series are randomly masked as missing data, where the maximum temporal length $l_t$ and feature dimension $l_c$ are set to 10 and 5, respectively.
          % This setting simulates outages or abnormal periods where data is entirely unavailable for certain durations. In this scenario, random contiguous blocks of varying lengths are masked as missing, simulating outages or abnormal periods where data is entirely unavailable for certain durations.
\end{itemize}

% These missing values scenarios reflect real-world challenges in multivariate time series forecasting with missing values, where information can be missing either sporadically or in larger chunks.

\begin{table}[!t]
    \caption{Detailed dataset descriptions.}
    \centering
    % \begin{center}
        \begin{tabular}{c|c|c|c|c}
            \toprule
            \textbf{Datasets} & \textbf{ \# Dims.} & \textbf{\# Points} & \textbf{Interval} & \textbf{Information} \\
            \hline
            ETTh1, ETTh2      & 7                  & 17420              & 1 hour            & Electricity          \\
            ETTm1, ETTm2      & 7                  & 69680              & 15 mins           & Electricity          \\
            % ECL               & 321                & 26304              & 1 hour            & Electricity          \\
            Weather           & 21                 & 36761              & 10 mins           & Weather              \\
            % Traffic           & 862                & 17451              & 1 hour            & Transportation       \\
            Exchange          & 8                  & 7588               & 1 day             & Economy              \\
            \bottomrule
            % \multicolumn{4}{l}{$^{\mathrm{a}}$Sample of a Table footnote.}
        \end{tabular}
        \label{tb1:dataset}
    % \end{center}
\end{table}

\begin{table}[t] 
    \centering
    \caption{Hyperparameter setting in our experiments. We keep the hyperparameters consistent across all datasets and missing-data scenarios.}
    \label{tab:hyperparams}
    \begin{tabular}{l|c}
        \toprule
        \textbf{Hyperparameters}      & \textbf{Values} \\
        \hline
        Hidden size                   & 256             \\
        Day-of-week embedding size    & 8               \\
        Hour-of-day embedding size    & 8               \\
        Dropout                       & 0.1             \\
        % Optimizer                     & Adam            \\
        Learning rate                 & 0.001           \\
        $\lambda$ (balance weight) & 0.2             \\
        Batch size                    & 512             \\
        Epochs                        & 100             \\
        \bottomrule
    \end{tabular}
\end{table}

\noindent{\textbf{Comparison Methods}:
To evaluate the effectiveness of our CoIFNet, we compare it with a wide spectrum of state-of-the-art approaches, including imputation-based methods, vanilla MTSF methods, and end-to-end methods. To ensure a fair comparison, we use mean imputation to initialize all missing values for the baseline methods.
% For consistency, all baseline implementations involve filling missing values with the mean of available data. The baselines are categorized as follows:

\begin{itemize}

    \item \textbf{Traditional MTSF Methods}:
          These methods apply masks in the loss function to reduce the impact of missing data on forecasting performance.
          % These models handle the missing values problem by masking the missing target values directly in the objective function. This allows the models to focus on learning from observed data while ignoring the influence of missing targets during training. Models in this category include:
          \begin{itemize}
              \item \textbf{STID\cite{shao2022spatial}:} A spatial-temporal interaction decomposition model that captures complex interactions in multivariate time series data.
              \item \textbf{DLinear\cite{zeng2023dlinear}:} A linear forecasting model optimized for computational efficiency and scalability in handling large datasets.
              \item \textbf{AGCRN\cite{bai2020adaptive}:} Combineing graph convolutional networks with recurrent architectures to model spatio-temporal dynamics effectively.
              \item \textbf{MTGNN\cite{wu2020connecting}:} Leveraging graph neural networks to model multivariate time series, focusing on spatial correlations and temporal patterns.
              \item \textbf{iTransformer\cite{liu2024itransformer}:} A transformer architecture based model that claims the capability to perform multiple tasks, including both imputation and forecasting, thereby providing a versatile solution for time series analysis.
              \item \textbf{GPT4TS\cite{zhou2023one}:} A large-scale generative pre-trained transformer model that emphasizes its ability to handle a variety of tasks such as imputation and forecasting, highlighting its adaptability and robustness in complex time series scenarios.
          \end{itemize}

    \item \textbf{Imputation-based Methods}:
          These methods treat the lookback sequence and forecast horizon as a unified sequence, with the entire forecast horizon set as missing during evaluation.
          \begin{itemize}
              \item \textbf{BRITS\cite{cao2018brits}:} Utilizing bidirectional recurrent neural networks to impute missing values, thereby enhancing the model's ability to learn temporal dependencies from incomplete data.
              \item \textbf{SAITS\cite{du2023saits}:} Employing self-attention mechanisms to reconstruct missing values, leveraging temporal context and observed patterns for imputation.
          \end{itemize}

    \item \textbf{End-to-End Methods}:
          % Specifically designed to address the challenges posed by incomplete observations in time series data, these models integrate missing values handling directly into their architecture. We include:
          \begin{itemize}
              \item \textbf{BiTGraph\cite{chen2024biased}:} The previous state-of-the-art end-to-end framework that jointly learns the temporal dependencies and spatial structure from time series with missing data.
                    % A model that leverages graph neural networks to effectively manage and predict in scenarios with partial observations.
          \end{itemize}
\end{itemize}

% Additionally, we compare our CoIFNet with the recently proposed BiTGraph—the current state-of-the-art end-to-end framework for MTSF with missing values.  

\begin{table*}[!t]
    \caption{Comparison with state-of-the-arts on six real-world datasets under the \textbf{{Point}} missing rates  of 0.3 and 0.6.}
    % \textbf{Bold} indicates the best result, and \underline{underline} indicates the second-best result. The last column AVG shows the average MSE and MAE results across all datasets.}
    \centering
    % \begin{center}
    \setlength{\tabcolsep}{3pt}
    \resizebox{\textwidth}{!}{
        \begin{tabular}{c|cc|cc|cc|cc|cc|cc|cc}
            \toprule
        
            \multicolumn{15}{c}{\textbf{Point}: $r=0.3$}                                                                                                                                                                                                                                                                                                                                                                                                      \\
            \hline
            Dataset          & \multicolumn{2}{c|}{ETTh1} & \multicolumn{2}{c|}{ETTh2} & \multicolumn{2}{c|}{ETTm1} & \multicolumn{2}{c|}{ETTm2} & \multicolumn{2}{c|}{Weather} & \multicolumn{2}{c|}{Exchange Rate} & \multicolumn{2}{c}{\textbf{AVG}}                                                                                                                                                                                                       \\
            Metric           & MAE                        & MSE                        & MAE                        & MSE                        & MAE                          & MSE                                & MAE                       & MSE                       & MAE                       & MSE                       & MAE                       & MSE                       & MAE                       & MSE                       \\
            \hline
            BRITS            & 0.814$_{\pm.01}$           & 1.207$_{\pm.02}$           & 1.156$_{\pm.01}$           & 2.361$_{\pm.06}$           & 0.751$_{\pm.02}$             & 1.027$_{\pm.02}$                   & 1.108$_{\pm.14}$          & 2.086$_{\pm.59}$          & 0.429$_{\pm.06}$          & 0.410$_{\pm.08}$          & 1.123$_{\pm.17}$          & 1.883$_{\pm.53}$          & 0.897$_{\pm.07}$          & 1.496$_{\pm.22}$          \\
            SAITS            & 0.848$_{\pm.00}$           & 1.271$_{\pm.01}$           & 1.188$_{\pm.01}$           & 2.454$_{\pm.02}$           & 0.790$_{\pm.00}$             & 1.098$_{\pm.01}$                   & 1.356$_{\pm.00}$          & 3.120$_{\pm.01}$          & 0.590$_{\pm.00}$          & 0.615$_{\pm.00}$          & 1.400$_{\pm.04}$          & 2.934$_{\pm.12}$          & 1.029$_{\pm.01}$          & 1.915$_{\pm.03}$          \\
            % \hline
            STID             & 0.469$_{\pm.00}$           & 0.494$_{\pm.00}$           & 0.369$_{\pm.01}$           & 0.297$_{\pm.01}$           & 0.374$_{\pm.00}$             & 0.342$_{\pm.00}$                   & 0.316$_{\pm.00}$          & 0.233$_{\pm.00}$          & 0.209$_{\pm.00}$          & 0.160$_{\pm.00}$          & 0.301$_{\pm.01}$          & 0.166$_{\pm.01}$          & 0.340$_{\pm.00}$          & 0.282$_{\pm.00}$          \\
            MTGNN            & 0.498$_{\pm.00}$           & 0.521$_{\pm.00}$           & 0.443$_{\pm.01}$           & 0.407$_{\pm.03}$           & 0.417$_{\pm.01}$             & 0.400$_{\pm.00}$                   & 0.333$_{\pm.00}$          & 0.238$_{\pm.01}$          & 0.226$_{\pm.01}$          & 0.179$_{\pm.01}$          & 0.329$_{\pm.00}$          & 0.196$_{\pm.00}$          & 0.374$_{\pm.01}$          & 0.323$_{\pm.01}$          \\
            AGCRN            & 0.651$_{\pm.02}$           & 0.846$_{\pm.03}$           & 0.589$_{\pm.04}$           & 0.775$_{\pm.10}$           & 0.498$_{\pm.00}$             & 0.515$_{\pm.02}$                   & 0.378$_{\pm.01}$          & 0.321$_{\pm.03}$          & 0.217$_{\pm.00}$          & 0.170$_{\pm.00}$          & 0.473$_{\pm.02}$          & 0.406$_{\pm.03}$          & 0.467$_{\pm.02}$          & 0.506$_{\pm.04}$          \\
            DLinear          & 0.463$_{\pm.00}$           & 0.490$_{\pm.00}$           & 0.365$_{\pm.00}$           & 0.287$_{\pm.00}$           & 0.387$_{\pm.00}$             & 0.369$_{\pm.00}$                   & 0.313$_{\pm.00}$          & 0.226$_{\pm.00}$          & 0.261$_{\pm.00}$          & 0.212$_{\pm.00}$          & 0.267$_{\pm.00}$          & 0.130$_{\pm.00}$          & 0.343$_{\pm.00}$          & 0.286$_{\pm.00}$          \\
            PatchTST         & 0.475$_{\pm.01}$           & 0.500$_{\pm.01}$           & 0.369$_{\pm.00}$           & 0.300$_{\pm.01}$           & 0.386$_{\pm.00}$             & 0.366$_{\pm.00}$                   & 0.307$_{\pm.01}$          & 0.224$_{\pm.01}$          & 0.238$_{\pm.00}$          & 0.190$_{\pm.00}$          & 0.259$_{\pm.00}$          & 0.128$_{\pm.01}$          & 0.339$_{\pm.00}$          & 0.285$_{\pm.01}$          \\
            iTransformer     & 0.465$_{\pm.00}$           & 0.488$_{\pm.00}$           & 0.362$_{\pm.00}$           & 0.288$_{\pm.00}$           & 0.382$_{\pm.00}$             & 0.356$_{\pm.00}$                   & 0.308$_{\pm.00}$          & 0.222$_{\pm.00}$          & 0.233$_{\pm.00}$          & 0.185$_{\pm.00}$          & 0.276$_{\pm.01}$          & 0.140$_{\pm.01}$          & 0.338$_{\pm.00}$          & 0.280$_{\pm.00}$          \\
            GPT4TS           & 0.481$_{\pm.00}$           & 0.506$_{\pm.01}$           & 0.398$_{\pm.00}$           & 0.339$_{\pm.01}$           & 0.385$_{\pm.00}$             & 0.363$_{\pm.00}$                   & 0.294$_{\pm.00}$          & 0.213$_{\pm.00}$          & 0.232$_{\pm.00}$          & 0.189$_{\pm.00}$          & 0.299$_{\pm.01}$          & 0.167$_{\pm.01}$          & 0.348$_{\pm.00}$          & 0.296$_{\pm.01}$          \\
            % \midrule
            BiTGraph         & 0.496$_{\pm.01}$           & 0.520$_{\pm.01}$           & 0.408$_{\pm.01}$           & 0.364$_{\pm.03}$           & 0.411$_{\pm.00}$             & 0.391$_{\pm.00}$                   & 0.323$_{\pm.01}$          & 0.230$_{\pm.01}$          & 0.222$_{\pm.00}$          & 0.174$_{\pm.00}$          & 0.320$_{\pm.02}$          & 0.188$_{\pm.03}$          & 0.363$_{\pm.01}$          & 0.311$_{\pm.01}$          \\
            \hline
            \rowcolor{green!10}\textbf{CoIFNet} & \textbf{0.452$_{\pm.00}$}  & \textbf{0.481$_{\pm.00}$}  & \textbf{0.315$_{\pm.00}$}  & \textbf{0.234$_{\pm.00}$}  & \textbf{0.351$_{\pm.00}$}    & \textbf{0.318$_{\pm.00}$}          & \textbf{0.252$_{\pm.00}$} & \textbf{0.172$_{\pm.00}$} & \textbf{0.195$_{\pm.00}$} & \textbf{0.153$_{\pm.00}$} & \textbf{0.209$_{\pm.00}$} & \textbf{0.088$_{\pm.00}$} & \textbf{0.295$_{\pm.00}$} & \textbf{0.241$_{\pm.00}$} \\
            \hline \hline

            \multicolumn{15}{c}{\textbf{Point}: $r=0.6$}                                                                                                                                                                                                                                                                                                                                                                                                      \\
            \hline
            Dataset          & \multicolumn{2}{c|}{ETTh1} & \multicolumn{2}{c|}{ETTh2} & \multicolumn{2}{c|}{ETTm1} & \multicolumn{2}{c|}{ETTm2} & \multicolumn{2}{c|}{Weather} & \multicolumn{2}{c|}{Exchange Rate} & \multicolumn{2}{c}{\textbf{AVG}}                                                                                                                                                                                                       \\
            Metric           & MAE                        & MSE                        & MAE                        & MSE                        & MAE                          & MSE                                & MAE                       & MSE                       & MAE                       & MSE                       & MAE                       & MSE                       & MAE                       & MSE                       \\
            \hline
            BRITS            & 0.823$_{\pm.02}$           & 1.214$_{\pm.04}$           & 1.001$_{\pm.11}$           & 1.759$_{\pm.46}$           & 0.795$_{\pm.01}$             & 1.060$_{\pm.02}$                   & 1.014$_{\pm.05}$          & 1.730$_{\pm.11}$          & 0.405$_{\pm.01}$          & 0.377$_{\pm.02}$          & 1.199$_{\pm.08}$          & 2.099$_{\pm.22}$          & 0.873$_{\pm.04}$          & 1.373$_{\pm.15}$          \\
            SAITS            & 0.842$_{\pm.01}$           & 1.272$_{\pm.01}$           & 1.187$_{\pm.01}$           & 2.455$_{\pm.03}$           & 0.787$_{\pm.00}$             & 1.090$_{\pm.01}$                   & 1.356$_{\pm.00}$          & 3.118$_{\pm.01}$          & 0.598$_{\pm.00}$          & 0.624$_{\pm.00}$          & 1.400$_{\pm.04}$          & 2.938$_{\pm.15}$          & 1.028$_{\pm.01}$          & 1.916$_{\pm.04}$          \\
            % \hline
            STID             & 0.497$_{\pm.01}$           & 0.537$_{\pm.00}$           & 0.415$_{\pm.00}$           & 0.363$_{\pm.00}$           & 0.395$_{\pm.00}$             & 0.370$_{\pm.00}$                   & 0.359$_{\pm.01}$          & 0.304$_{\pm.01}$          & 0.222$_{\pm.00}$          & 0.170$_{\pm.00}$          & 0.337$_{\pm.02}$          & 0.207$_{\pm.03}$          & 0.371$_{\pm.01}$          & 0.325$_{\pm.01}$          \\
            MTGNN            & 0.521$_{\pm.01}$           & 0.558$_{\pm.01}$           & 0.447$_{\pm.00}$           & 0.406$_{\pm.01}$           & 0.433$_{\pm.01}$             & 0.425$_{\pm.02}$                   & 0.363$_{\pm.02}$          & 0.279$_{\pm.03}$          & 0.236$_{\pm.01}$          & 0.187$_{\pm.01}$          & 0.376$_{\pm.01}$          & 0.254$_{\pm.02}$          & 0.396$_{\pm.01}$          & 0.352$_{\pm.01}$          \\
            AGCRN            & 0.677$_{\pm.02}$           & 0.889$_{\pm.02}$           & 0.578$_{\pm.01}$           & 0.712$_{\pm.01}$           & 0.496$_{\pm.01}$             & 0.535$_{\pm.03}$                   & 0.408$_{\pm.02}$          & 0.348$_{\pm.04}$          & 0.229$_{\pm.01}$          & 0.178$_{\pm.01}$          & 0.541$_{\pm.04}$          & 0.491$_{\pm.05}$          & 0.488$_{\pm.02}$          & 0.525$_{\pm.03}$          \\
            DLinear          & 0.500$_{\pm.00}$           & 0.542$_{\pm.00}$           & 0.403$_{\pm.00}$           & 0.339$_{\pm.00}$           & 0.424$_{\pm.00}$             & 0.421$_{\pm.00}$                   & 0.355$_{\pm.00}$          & 0.274$_{\pm.01}$          & 0.281$_{\pm.00}$          & 0.224$_{\pm.00}$          & 0.306$_{\pm.01}$          & 0.167$_{\pm.01}$          & 0.378$_{\pm.00}$          & 0.328$_{\pm.00}$          \\
            PatchTST         & 0.501$_{\pm.00}$           & 0.538$_{\pm.01}$           & 0.372$_{\pm.01}$           & 0.304$_{\pm.02}$           & 0.419$_{\pm.01}$             & 0.414$_{\pm.01}$                   & 0.315$_{\pm.00}$          & 0.234$_{\pm.00}$          & 0.250$_{\pm.00}$          & 0.203$_{\pm.00}$          & 0.274$_{\pm.01}$          & 0.145$_{\pm.01}$          & 0.355$_{\pm.01}$          & 0.306$_{\pm.01}$          \\
            iTransformer     & 0.491$_{\pm.00}$           & 0.523$_{\pm.00}$           & 0.398$_{\pm.00}$           & 0.335$_{\pm.00}$           & 0.410$_{\pm.00}$             & 0.393$_{\pm.00}$                   & 0.352$_{\pm.00}$          & 0.270$_{\pm.00}$          & 0.234$_{\pm.00}$          & 0.189$_{\pm.00}$          & 0.312$_{\pm.01}$          & 0.175$_{\pm.01}$          & 0.366$_{\pm.00}$          & 0.314$_{\pm.00}$          \\
            GPT4TS           & 0.524$_{\pm.01}$           & 0.568$_{\pm.01}$           & 0.436$_{\pm.02}$           & 0.414$_{\pm.05}$           & 0.406$_{\pm.00}$             & 0.391$_{\pm.00}$                   & 0.331$_{\pm.02}$          & 0.272$_{\pm.04}$          & 0.255$_{\pm.00}$          & 0.212$_{\pm.00}$          & 0.408$_{\pm.03}$          & 0.331$_{\pm.05}$          & 0.393$_{\pm.01}$          & 0.365$_{\pm.02}$          \\
            % \midrule
            BiTGraph         & 0.515$_{\pm.01}$           & 0.548$_{\pm.01}$           & 0.444$_{\pm.01}$           & 0.419$_{\pm.04}$           & 0.421$_{\pm.00}$             & 0.408$_{\pm.01}$                   & 0.339$_{\pm.01}$          & 0.256$_{\pm.02}$          & 0.226$_{\pm.00}$          & 0.179$_{\pm.01}$          & 0.319$_{\pm.00}$          & 0.184$_{\pm.00}$          & 0.377$_{\pm.01}$          & 0.332$_{\pm.01}$          \\
            \hline
           \rowcolor{green!10} \textbf{CoIFNet} & \textbf{0.467$_{\pm.00}$}  & \textbf{0.499$_{\pm.00}$}  & \textbf{0.323$_{\pm.00}$}  & \textbf{0.246$_{\pm.00}$}  & \textbf{0.363$_{\pm.00}$}    & \textbf{0.337$_{\pm.01}$}          & \textbf{0.256$_{\pm.00}$} & \textbf{0.176$_{\pm.00}$} & \textbf{0.200$_{\pm.00}$} & \textbf{0.157$_{\pm.00}$} & \textbf{0.216$_{\pm.00}$} & \textbf{0.093$_{\pm.00}$} & \textbf{0.304$_{\pm.00}$} & \textbf{0.251$_{\pm.00}$} \\
            \bottomrule
        \end{tabular}
    }
    \label{tb:Point}
    % \end{center}
\end{table*}

\noindent{\textbf{Implementation Details}:
% For all experiments, we use a consistent configuration across datasets and models. 
In our experiments, each model is trained using the historical lookback window of length $L=96$ to predict the future forecasting window of length $H=96$.
Table~\ref{tab:hyperparams} presents the hyperparameter setting in our experiments. Note that, we keep the hyperparameters consistent across all datasets and missing-data scenarios. In other words, there is no cherry-picking of parameters for each setting.
Besides, the missing rates for the point/block missing pattern are set to \{0.3, 0.6\}.
% Table~\ref{tab:hyperparams} presents the hyperparameter settings used in our CoIFNet model. 
% we maintain consistent hyperparameter settings across all datasets and value missing patterns.
% The hyperparameters were determined through preliminary experiments on the validation sets. 
We employ early stopping with a patience of 10 epochs to prevent overfitting and select the model with the best validation performance.
% All models are implemented in PyTorch and trained/tested on a Linux machine with one NVIDIA RTX 4090 GPU. 
For each experiment, we conduct three independent runs with different random seeds and report the average MAE and MSE results.
% to ensure statistical robustness.
% We evaluate the performance using two metrics of Mean Squared Error (MSE) and Mean Absolute Error (MAE). % To ensure reproducibility, 
We have made our code publicly available at \textcolor{magenta}{\url{https://github.com/KaiTang-eng/CoIFNet}}.

\begin{table*}[htbp]
    \caption{Comparison with state-of-the-arts on six real-world datasets under the \textbf{{Block}}  missing rates of 0.3 and 0.6. }
    \centering
    % \vspace{-3pt}
    % \begin{center}
    \setlength{\tabcolsep}{3pt}
    \resizebox{\textwidth}{!}{
        \begin{tabular}{c|cc|cc|cc|cc|cc|cc|cc}
            \toprule
            \multicolumn{15}{c}{\textbf{Block}: $r=0.3$}                                                                                     \\
            \hline
            Dataset          & \multicolumn{2}{c|}{ETTh1} & \multicolumn{2}{c|}{ETTh2} & \multicolumn{2}{c|}{ETTm1} & \multicolumn{2}{c|}{ETTm2} & \multicolumn{2}{c|}{Weather} & \multicolumn{2}{c|}{Exchange Rate} & \multicolumn{2}{c}{\textbf{AVG}}                                                                                                                                                                                                       \\
            Metric           & MAE                        & MSE                        & MAE                        & MSE                        & MAE                          & MSE                                & MAE                       & MSE                       & MAE                       & MSE                       & MAE                       & MSE                       & MAE                       & MSE                       \\
            \hline
            BRITS            & 0.794$_{\pm.02}$           & 1.173$_{\pm.05}$           & 1.115$_{\pm.08}$           & 2.152$_{\pm.28}$           & 0.757$_{\pm.01}$             & 1.034$_{\pm.03}$                   & 1.231$_{\pm.04}$          & 2.550$_{\pm.28}$          & 0.422$_{\pm.08}$          & 0.402$_{\pm.10}$          & 1.074$_{\pm.05}$          & 1.690$_{\pm.15}$          & 0.899$_{\pm.05}$          & 1.500$_{\pm.15}$          \\
            SAITS            & 0.850$_{\pm.00}$           & 1.283$_{\pm.01}$           & 1.188$_{\pm.00}$           & 2.454$_{\pm.01}$           & 0.789$_{\pm.00}$             & 1.096$_{\pm.01}$                   & 1.353$_{\pm.00}$          & 3.107$_{\pm.03}$          & 0.594$_{\pm.00}$          & 0.618$_{\pm.00}$          & 1.395$_{\pm.03}$          & 2.907$_{\pm.11}$          & 1.028$_{\pm.01}$          & 1.911$_{\pm.03}$          \\
            % \hline
            STID             & 0.494$_{\pm.01}$           & 0.539$_{\pm.02}$           & 0.441$_{\pm.01}$           & 0.436$_{\pm.03}$           & 0.391$_{\pm.00}$             & 0.372$_{\pm.01}$                   & 0.361$_{\pm.02}$          & 0.319$_{\pm.03}$          & 0.220$_{\pm.00}$          & 0.169$_{\pm.00}$          & 0.340$_{\pm.00}$          & 0.223$_{\pm.00}$          & 0.375$_{\pm.01}$          & 0.343$_{\pm.01}$          \\
            MTGNN            & 0.518$_{\pm.01}$           & 0.559$_{\pm.01}$           & 0.486$_{\pm.00}$           & 0.486$_{\pm.02}$           & 0.432$_{\pm.00}$             & 0.423$_{\pm.00}$                   & 0.349$_{\pm.00}$          & 0.258$_{\pm.00}$          & 0.231$_{\pm.00}$          & 0.183$_{\pm.00}$          & 0.390$_{\pm.02}$          & 0.281$_{\pm.04}$          & 0.401$_{\pm.01}$          & 0.365$_{\pm.01}$          \\
            AGCRN            & 0.660$_{\pm.01}$           & 0.867$_{\pm.02}$           & 0.604$_{\pm.02}$           & 0.828$_{\pm.04}$           & 0.515$_{\pm.01}$             & 0.563$_{\pm.02}$                   & 0.434$_{\pm.04}$          & 0.417$_{\pm.07}$          & 0.238$_{\pm.00}$          & 0.187$_{\pm.01}$          & 0.535$_{\pm.01}$          & 0.505$_{\pm.01}$          & 0.497$_{\pm.02}$          & 0.561$_{\pm.03}$          \\
            DLinear          & 0.491$_{\pm.00}$           & 0.542$_{\pm.01}$           & 0.450$_{\pm.01}$           & 0.438$_{\pm.02}$           & 0.425$_{\pm.00}$             & 0.424$_{\pm.00}$                   & 0.417$_{\pm.01}$          & 0.395$_{\pm.02}$          & 0.289$_{\pm.00}$          & 0.232$_{\pm.00}$          & 0.376$_{\pm.01}$          & 0.254$_{\pm.02}$          & 0.408$_{\pm.00}$          & 0.381$_{\pm.01}$          \\
            PatchTST         & 0.492$_{\pm.00}$           & 0.528$_{\pm.01}$           & 0.417$_{\pm.01}$           & 0.387$_{\pm.01}$           & 0.416$_{\pm.00}$             & 0.407$_{\pm.00}$                   & 0.360$_{\pm.01}$          & 0.319$_{\pm.01}$          & 0.259$_{\pm.00}$          & 0.206$_{\pm.00}$          & 0.343$_{\pm.02}$          & 0.239$_{\pm.04}$          & 0.381$_{\pm.01}$          & 0.348$_{\pm.01}$          \\
            iTransformer     & 0.498$_{\pm.00}$           & 0.540$_{\pm.01}$           & 0.465$_{\pm.01}$           & 0.461$_{\pm.02}$           & 0.423$_{\pm.01}$             & 0.413$_{\pm.01}$                   & 0.416$_{\pm.01}$          & 0.381$_{\pm.01}$          & 0.256$_{\pm.00}$          & 0.203$_{\pm.00}$          & 0.396$_{\pm.02}$          & 0.296$_{\pm.04}$          & 0.409$_{\pm.01}$          & 0.382$_{\pm.01}$          \\
            GPT4TS           & 0.496$_{\pm.00}$           & 0.544$_{\pm.01}$           & 0.480$_{\pm.02}$           & 0.514$_{\pm.05}$           & 0.407$_{\pm.00}$             & 0.395$_{\pm.00}$                   & 0.390$_{\pm.01}$          & 0.400$_{\pm.02}$          & 0.254$_{\pm.00}$          & 0.204$_{\pm.00}$          & 0.359$_{\pm.01}$          & 0.275$_{\pm.01}$          & 0.398$_{\pm.01}$          & 0.389$_{\pm.02}$          \\
            % \midrule
            BiTGraph         & 0.519$_{\pm.01}$           & 0.558$_{\pm.01}$           & 0.430$_{\pm.01}$           & 0.384$_{\pm.02}$           & 0.420$_{\pm.00}$             & 0.408$_{\pm.00}$                   & 0.336$_{\pm.02}$          & 0.248$_{\pm.02}$          & 0.229$_{\pm.00}$          & 0.181$_{\pm.01}$          & 0.381$_{\pm.04}$          & 0.266$_{\pm.07}$          & 0.386$_{\pm.01}$          & 0.341$_{\pm.02}$          \\
            \hline
             \rowcolor{green!10} \textbf{CoIFNet} & \textbf{0.467$_{\pm.00}$}  & \textbf{0.508$_{\pm.01}$}  & \textbf{0.319$_{\pm.00}$}  & \textbf{0.240$_{\pm.00}$}  & \textbf{0.369$_{\pm.00}$}    & \textbf{0.343$_{\pm.00}$}          & \textbf{0.257$_{\pm.00}$} & \textbf{0.179$_{\pm.00}$} & \textbf{0.201$_{\pm.00}$} & \textbf{0.158$_{\pm.00}$} & \textbf{0.218$_{\pm.00}$} & \textbf{0.097$_{\pm.00}$} & \textbf{0.305$_{\pm.00}$} & \textbf{0.254$_{\pm.00}$} \\
            \hline \hline
            \multicolumn{15}{c}{\textbf{Block}: $r=0.6$}                                                                                                                                                                                                                                                                                                                                                                                                    \\
            \hline
            Dataset          & \multicolumn{2}{c|}{ETTh1} & \multicolumn{2}{c|}{ETTh2} & \multicolumn{2}{c|}{ETTm1} & \multicolumn{2}{c|}{ETTm2} & \multicolumn{2}{c|}{Weather} & \multicolumn{2}{c|}{Exchange Rate} & \multicolumn{2}{c}{\textbf{AVG}}                                                                                                                                                                                                       \\
            Metric           & MAE                        & MSE                        & MAE                        & MSE                        & MAE                          & MSE                                & MAE                       & MSE                       & MAE                       & MSE                       & MAE                       & MSE                       & MAE                       & MSE                       \\
            \hline
            BRITS            & 0.796$_{\pm.04}$           & 1.172$_{\pm.07}$           & 1.031$_{\pm.02}$           & 1.813$_{\pm.13}$           & 0.774$_{\pm.03}$             & 1.042$_{\pm.04}$                   & 1.199$_{\pm.03}$          & 2.359$_{\pm.25}$          & 0.386$_{\pm.04}$          & 0.348$_{\pm.04}$          & 1.077$_{\pm.09}$          & 1.728$_{\pm.29}$          & 0.877$_{\pm.04}$          & 1.410$_{\pm.14}$          \\
            SAITS            & 0.844$_{\pm.00}$           & 1.265$_{\pm.00}$           & 1.183$_{\pm.01}$           & 2.436$_{\pm.03}$           & 0.788$_{\pm.00}$                          & 1.097$_{\pm.00}$                                & 1.362$_{\pm.00}$          & 3.145$_{\pm.01}$          & 0.600$_{\pm.00}$          & 0.628$_{\pm.00}$          & 1.399$_{\pm.04}$          & 2.922$_{\pm.13}$          & 2.398$_{\pm.51}$          & 3.233$_{\pm.53}$          \\
            % \hline
            STID             & 0.539$_{\pm.00}$           & 0.612$_{\pm.01}$           & 0.488$_{\pm.01}$           & 0.517$_{\pm.03}$           & 0.419$_{\pm.00}$             & 0.407$_{\pm.01}$                   & 0.389$_{\pm.01}$          & 0.354$_{\pm.02}$          & 0.234$_{\pm.00}$          & 0.181$_{\pm.00}$          & 0.400$_{\pm.02}$          & 0.307$_{\pm.03}$          & 0.412$_{\pm.01}$          & 0.396$_{\pm.02}$          \\
            MTGNN            & 0.552$_{\pm.01}$           & 0.625$_{\pm.02}$           & 0.522$_{\pm.01}$           & 0.562$_{\pm.02}$           & 0.471$_{\pm.03}$             & 0.485$_{\pm.05}$                   & 0.383$_{\pm.01}$          & 0.307$_{\pm.01}$          & 0.242$_{\pm.00}$          & 0.192$_{\pm.00}$          & 0.434$_{\pm.04}$          & 0.346$_{\pm.08}$          & 0.434$_{\pm.02}$          & 0.420$_{\pm.03}$          \\
            AGCRN            & 0.727$_{\pm.06}$           & 0.970$_{\pm.09}$           & 0.651$_{\pm.03}$           & 0.922$_{\pm.05}$           & 0.546$_{\pm.02}$             & 0.598$_{\pm.04}$                   & 0.478$_{\pm.02}$          & 0.458$_{\pm.03}$          & 0.250$_{\pm.00}$          & 0.197$_{\pm.00}$          & 0.598$_{\pm.02}$          & 0.604$_{\pm.03}$          & 0.542$_{\pm.03}$          & 0.625$_{\pm.04}$          \\
            DLinear          & 0.549$_{\pm.00}$           & 0.643$_{\pm.01}$           & 0.524$_{\pm.01}$           & 0.573$_{\pm.01}$           & 0.478$_{\pm.00}$             & 0.508$_{\pm.01}$                   & 0.506$_{\pm.00}$          & 0.545$_{\pm.01}$          & 0.314$_{\pm.00}$          & 0.253$_{\pm.00}$          & 0.450$_{\pm.02}$          & 0.363$_{\pm.04}$          & 0.470$_{\pm.01}$          & 0.481$_{\pm.01}$          \\
            PatchTST         & 0.537$_{\pm.00}$           & 0.605$_{\pm.02}$           & 0.391$_{\pm.01}$           & 0.328$_{\pm.02}$           & 0.467$_{\pm.00}$             & 0.473$_{\pm.01}$                   & 0.331$_{\pm.00}$          & 0.259$_{\pm.00}$          & 0.270$_{\pm.00}$          & 0.219$_{\pm.00}$          & 0.384$_{\pm.02}$          & 0.311$_{\pm.04}$          & 0.397$_{\pm.01}$          & 0.366$_{\pm.01}$          \\
            iTransformer     & 0.543$_{\pm.00}$           & 0.621$_{\pm.01}$           & 0.527$_{\pm.01}$           & 0.580$_{\pm.03}$           & 0.460$_{\pm.00}$             & 0.472$_{\pm.00}$                   & 0.383$_{\pm.00}$          & 0.320$_{\pm.01}$          & 0.249$_{\pm.00}$          & 0.204$_{\pm.00}$          & 0.432$_{\pm.03}$          & 0.360$_{\pm.05}$          & 0.432$_{\pm.01}$          & 0.426$_{\pm.02}$          \\
            GPT4TS           & 0.579$_{\pm.01}$           & 0.661$_{\pm.02}$           & 0.453$_{\pm.00}$           & 0.425$_{\pm.01}$           & 0.477$_{\pm.01}$             & 0.480$_{\pm.01}$                   & 0.375$_{\pm.01}$          & 0.318$_{\pm.02}$          & 0.277$_{\pm.00}$          & 0.236$_{\pm.01}$          & 0.513$_{\pm.02}$          & 0.522$_{\pm.04}$          & 0.446$_{\pm.01}$          & 0.441$_{\pm.02}$          \\
            % \midrule
            BiTGraph         & 0.543$_{\pm.02}$           & 0.608$_{\pm.03}$           & 0.448$_{\pm.01}$           & 0.417$_{\pm.01}$           & 0.442$_{\pm.01}$             & 0.439$_{\pm.02}$                   & 0.337$_{\pm.01}$          & 0.249$_{\pm.02}$          & 0.234$_{\pm.00}$          & 0.186$_{\pm.00}$          & 0.359$_{\pm.02}$          & 0.243$_{\pm.04}$          & 0.394$_{\pm.01}$          & 0.357$_{\pm.02}$          \\
            \hline
           \rowcolor{green!10} \textbf{CoIFNet} & \textbf{0.490$_{\pm.00}$}  & \textbf{0.552$_{\pm.01}$}  & \textbf{0.330$_{\pm.00}$}  & \textbf{0.253$_{\pm.00}$}  & \textbf{0.389$_{\pm.01}$}    & \textbf{0.377$_{\pm.02}$}          & \textbf{0.261$_{\pm.00}$} & \textbf{0.184$_{\pm.00}$} & \textbf{0.207$_{\pm.00}$} & \textbf{0.163$_{\pm.00}$} & \textbf{0.225$_{\pm.00}$} & \textbf{0.101$_{\pm.00}$} & \textbf{0.317$_{\pm.00}$} & \textbf{0.272$_{\pm.00}$} \\
            \bottomrule
        \end{tabular}
    }
    \label{tb:Block}
    % \end{center}
\end{table*}

\begin{table*}[!t]
	\setlength{\abovecaptionskip}{0pt}  % 标题上方间距
	\setlength{\belowcaptionskip}{0pt}  % 标题下方间距
	\caption{Ablation study of the devised components of CoIFNet on the ETTm2 and Weather datasets at the point/block missing rates of 0.3 and 0.6. 
     “w/o CTF/CVF” indicates the CTF/CVF component is replaced by a linear layer.}
	\begin{center}
    \resizebox{\textwidth}{!}{
		\begin{tabular}{c|l|cccc|cccc}
			\toprule
			\multicolumn{2}{c|}{\multirow{3}{*}{Setting}} & \multicolumn{4}{c|}{ETTm2}                          & \multicolumn{4}{c}{Weather}                                                                                                                                                                                                                                                                   \\
			\multicolumn{2}{c|}{}                         & \multicolumn{2}{c}{$r=0.3$}                         & \multicolumn{2}{c|}{$r=0.6$}      & \multicolumn{2}{c}{$r=0.3$}       & \multicolumn{2}{c}{$r=0.6$}                                                                                                                                                                                           \\
			\multicolumn{2}{c|}{}                         & MAE                                                 & MSE                               & MAE                               & MSE                               & MAE                               & MSE                               & MAE                               & MSE                                                                   \\
			\hline
             % & \circled{9} Predict $\hat{\mathbf{Y}}$ rather than  $[\hat{\mathbf{X}}, \hat{\mathbf{Y}}]$    & 0.263$_{\pm \text{.00}}$          & 0.189$_{\pm \text{.00}}$          & 0.271$_{\pm \text{.00}}$          & 0.200$_{\pm \text{.00}}$          & 0.212$_{\pm \text{.01}}$          & 0.169$_{\pm \text{.00}}$          & 0.218$_{\pm \text{.01}}$          & 0.174$_{\pm \text{.01}}$          \\
	       \multirow{8}{*}{\rotatebox{90}{\textbf{Point}} }       & CoIFNet     \cellcolor{green!10}   & \textbf{0.252$_{\pm \text{.00}}$} \cellcolor{green!10}& \textbf{0.172$_{\pm \text{.00}}$} \cellcolor{green!10}& \textbf{0.258$_{\pm \text{.00}}$} \cellcolor{green!10}& \textbf{0.179$_{\pm \text{.00}}$} \cellcolor{green!10}& \textbf{0.200$_{\pm \text{.00}}$} \cellcolor{green!10}& \textbf{0.157$_{\pm \text{.00}}$} \cellcolor{green!10}& \textbf{0.204$_{\pm \text{.00}}$} \cellcolor{green!10}& \cellcolor{green!10} \textbf{0.161$_{\pm \text{.00}}$} \\
             \cline{2-10}
			                                              & \circled{1} w/o $\mathbf{M}^x$                      & 0.254$_{\pm \text{.00}}$          & 0.174$_{\pm \text{.00}}$          & 0.260$_{\pm \text{.00}}$          & 0.181$_{\pm \text{.00}}$          & 0.200$_{\pm \text{.00}}$          & 0.157$_{\pm \text{.00}}$          & 0.206$_{\pm \text{.00}}$          & 0.163$_{\pm \text{.00}}$          \\
			                                              & \circled{2} w/o $E_{\tau}$                              & 0.253$_{\pm \text{.00}}$          & 0.173$_{\pm \text{.00}}$          & 0.262$_{\pm \text{.00}}$          & 0.183$_{\pm \text{.00}}$          & 0.203$_{\pm \text{.00}}$          & 0.159$_{\pm \text{.00}}$          & 0.208$_{\pm \text{.00}}$          & 0.164$_{\pm \text{.00}}$          \\
			                                              & \circled{3} w/o $\mathbf{M}^x$ \& $E_{\tau}$        & 0.256$_{\pm \text{.00}}$          & 0.175$_{\pm \text{.00}}$          & 0.265$_{\pm \text{.00}}$          & 0.186$_{\pm \text{.00}}$          & 0.201$_{\pm \text{.00}}$          & 0.158$_{\pm \text{.00}}$          & 0.209$_{\pm \text{.00}}$          & 0.164$_{\pm \text{.00}}$          \\
 & \circled{4} w/o CTF  & 0.260$_{\pm \text{.00}}$ & 0.183$_{\pm \text{.00}}$& 0.266$_{\pm \text{.00}}$ & 0.190$_{\pm \text{.00}}$& 0.204$_{\pm \text{.00}}$ & 0.160$_{\pm \text{.00}}$& 0.210$_{\pm \text{.00}}$ & 0.166$_{\pm \text{.00}}$ \\ 
 & \circled{5} w/o CVF  & 0.256$_{\pm \text{.00}}$ & 0.176$_{\pm \text{.00}}$& 0.264$_{\pm \text{.00}}$ & 0.185$_{\pm \text{.00}}$& 0.210$_{\pm \text{.00}}$ & 0.169$_{\pm \text{.01}}$& 0.214$_{\pm \text{.00}}$ & 0.169$_{\pm \text{.00}}$ \\ 
 & \circled{6} w/o CTF \& CVF  & 0.264$_{\pm \text{.00}}$ & 0.187$_{\pm \text{.00}}$& 0.271$_{\pm \text{.00}}$ & 0.194$_{\pm \text{.00}}$& 0.233$_{\pm \text{.00}}$ & 0.197$_{\pm \text{.00}}$& 0.236$_{\pm \text{.00}}$ & 0.198$_{\pm \text{.00}}$ \\ 
			                                              & \circled{7} w/o RevON                               & 0.374$_{\pm \text{.04}}$          & 0.389$_{\pm \text{.11}}$          & 0.403$_{\pm \text{.06}}$          & 0.445$_{\pm \text{.21}}$          & 0.206$_{\pm \text{.01}}$          & 0.156$_{\pm \text{.01}}$          & 0.213$_{\pm \text{.00}}$          & 0.161$_{\pm \text{.00}}$          \\
			                                              % & \quad \color{gray} \textit{v1}. Remove Affine Layer & 0.253$_{\pm \text{.00}}$          & 0.173$_{\pm \text{.00}}$          & 0.261$_{\pm \text{.00}}$          & 0.182$_{\pm \text{.00}}$          & 0.203$_{\pm \text{.00}}$          & 0.160$_{\pm \text{.00}}$          & 0.207$_{\pm \text{.00}}$          & 0.163$_{\pm \text{.00}}$          \\
                                                          % \cline{2-10}
			                                              & \circled{8} Replace RevON by RevIN \cite{kim2021reversible} & 0.369$_{\pm \text{.01}}$          & 0.316$_{\pm \text{.02}}$          & 0.473$_{\pm \text{.04}}$          & 0.514$_{\pm \text{.09}}$          & 0.217$_{\pm \text{.00}}$          & 0.163$_{\pm \text{.00}}$          & 0.242$_{\pm \text{.00}}$          & 0.189$_{\pm \text{.00}}$          \\
             &\circled{9} Predict $\hat{\mathbf{Y}}$ only, rather than  $[\hat{\mathbf{X}}, \hat{\mathbf{Y}}]$    & 0.263$_{\pm \text{.00}}$          & 0.189$_{\pm \text{.00}}$          & 0.271$_{\pm \text{.00}}$          & 0.200$_{\pm \text{.00}}$          & 0.212$_{\pm \text{.01}}$          & 0.169$_{\pm \text{.00}}$          & 0.218$_{\pm \text{.01}}$          & 0.174$_{\pm \text{.01}}$          \\
			% \hline
            % \hline
 % & \circled{9} Predict $\hat{\mathbf{Y}}$ rather than  $[\hat{\mathbf{X}}, \hat{\mathbf{Y}}]$         & 0.269$_{\pm \text{.00}}$          & 0.196$_{\pm \text{.00}}$          & 0.271$_{\pm \text{.00}}$          & 0.196$_{\pm \text{.00}}$          & 0.224$_{\pm \text{.01}}$          & 0.178$_{\pm \text{.01}}$          & 0.227$_{\pm \text{.01}}$          & 0.182$_{\pm \text{.01}}$          \\
 \hline
			\multirow{8}{*}{\rotatebox{90}{\textbf{Block}} }       & CoIFNet   \cellcolor{green!10}                              & \textbf{0.258$_{\pm \text{.00}}$} \cellcolor{green!10} & \cellcolor{green!10} \textbf{0.179$_{\pm \text{.00}}$} & \cellcolor{green!10} \textbf{0.263$_{\pm \text{.00}}$} &  \cellcolor{green!10} \textbf{0.186$_{\pm \text{.00}}$} & \cellcolor{green!10} \textbf{0.203$_{\pm \text{.00}}$} & \cellcolor{green!10} \textbf{0.159$_{\pm \text{.00}}$} & \cellcolor{green!10} \textbf{0.213$_{\pm \text{.00}}$} & \cellcolor{green!10} \textbf{0.168$_{\pm \text{.00}}$} \\
             \cline{2-10}
			                                              & \circled{1} w/o $\mathbf{M}^x$                      & 0.262$_{\pm \text{.00}}$          & 0.185$_{\pm \text{.00}}$          & 0.267$_{\pm \text{.00}}$          & 0.191$_{\pm \text{.00}}$          & 0.207$_{\pm \text{.00}}$          & 0.163$_{\pm \text{.00}}$          & 0.217$_{\pm \text{.00}}$          & 0.173$_{\pm \text{.00}}$          \\
			                                              & \circled{2} w/o $E_{\tau}$                              & 0.261$_{\pm \text{.00}}$          & 0.182$_{\pm \text{.00}}$          & 0.269$_{\pm \text{.00}}$          & 0.192$_{\pm \text{.00}}$          & 0.208$_{\pm \text{.00}}$          & 0.163$_{\pm \text{.00}}$          & 0.218$_{\pm \text{.00}}$          & 0.173$_{\pm \text{.00}}$          \\
			                                              & \circled{3} w/o $\mathbf{M}^x$ \& $E_{\tau}$        & 0.263$_{\pm \text{.00}}$          & 0.184$_{\pm \text{.00}}$          & 0.271$_{\pm \text{.00}}$          & 0.195$_{\pm \text{.00}}$          & 0.210$_{\pm \text{.00}}$          & 0.163$_{\pm \text{.00}}$          & 0.217$_{\pm \text{.00}}$          & 0.171$_{\pm \text{.00}}$          \\
 & \circled{4} w/o CTF  & 0.265$_{\pm \text{.00}}$ & 0.190$_{\pm \text{.01}}$& 0.268$_{\pm \text{.00}}$ & 0.191$_{\pm \text{.00}}$& 0.212$_{\pm \text{.00}}$ & 0.167$_{\pm \text{.00}}$& 0.221$_{\pm \text{.00}}$ & 0.177$_{\pm \text{.00}}$ \\ 
 & \circled{5} w/o CVF  & 0.263$_{\pm \text{.00}}$ & 0.186$_{\pm \text{.00}}$& 0.270$_{\pm \text{.00}}$ & 0.194$_{\pm \text{.00}}$& 0.209$_{\pm \text{.00}}$ & 0.164$_{\pm \text{.00}}$& 0.219$_{\pm \text{.00}}$ & 0.173$_{\pm \text{.00}}$ \\ 
 & \circled{6} w/o CTF \& CVF  & 0.268$_{\pm \text{.00}}$ & 0.193$_{\pm \text{.00}}$& 0.277$_{\pm \text{.00}}$ & 0.201$_{\pm \text{.00}}$& 0.236$_{\pm \text{.00}}$ & 0.196$_{\pm \text{.00}}$& 0.240$_{\pm \text{.00}}$ & 0.200$_{\pm \text{.00}}$ \\ 			                              & \circled{7} w/o RevON                               & 0.405$_{\pm \text{.06}}$          & 0.485$_{\pm \text{.20}}$          & 0.441$_{\pm \text{.02}}$          & 0.471$_{\pm \text{.06}}$          & 0.220$_{\pm \text{.01}}$          & 0.168$_{\pm \text{.00}}$          & 0.224$_{\pm \text{.00}}$          & 0.173$_{\pm \text{.01}}$          \\
			                                              % & \quad \color{gray} \textit{v1}. Remove Affine Layer & 0.260$_{\pm \text{.00}}$          & 0.182$_{\pm \text{.00}}$          & 0.270$_{\pm \text{.00}}$          & 0.193$_{\pm \text{.00}}$          & 0.209$_{\pm \text{.00}}$          & 0.164$_{\pm \text{.00}}$          & 0.217$_{\pm \text{.00}}$          & 0.172$_{\pm \text{.00}}$          \\
			                                              &\circled{8} Replace RevON by RevIN \cite{kim2021reversible} & 0.372$_{\pm \text{.01}}$          & 0.334$_{\pm \text{.01}}$          & 0.498$_{\pm \text{.03}}$          & 0.567$_{\pm \text{.06}}$          & 0.227$_{\pm \text{.00}}$          & 0.168$_{\pm \text{.00}}$          & 0.240$_{\pm \text{.00}}$          & 0.181$_{\pm \text{.00}}$          \\
 & \circled{9} Predict $\hat{\mathbf{Y}}$ only, rather than  $[\hat{\mathbf{X}}, \hat{\mathbf{Y}}]$         & 0.269$_{\pm \text{.00}}$          & 0.196$_{\pm \text{.00}}$          & 0.271$_{\pm \text{.00}}$          & 0.196$_{\pm \text{.00}}$          & 0.224$_{\pm \text{.01}}$          & 0.178$_{\pm \text{.01}}$          & 0.227$_{\pm \text{.01}}$          & 0.182$_{\pm \text{.01}}$          \\
			\bottomrule
		\end{tabular}}
	\end{center}
	\label{tb:ablation}
\end{table*}

\begin{figure*}[htbp]
    \setlength{\abovecaptionskip}{3pt}  % 标题上方间距
    \setlength{\belowcaptionskip}{0.3pt}  % 标题下方间距
    \centerline{\includegraphics[width=\textwidth]{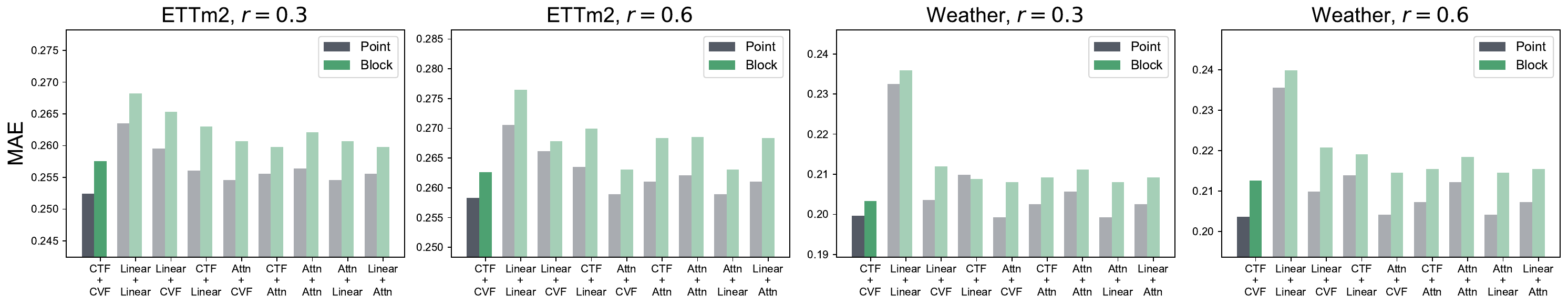}}
    % \centerline{\includegraphics[width=\columnwidth]{fig/ablationCF.pdf}}
    % \centerline{\includegraphics[width=0.95\columnwidth]{fig/ablation.pdf}}
    \caption{Ablation study of the CTF and CVF modules on ETTm2 and Weather datasets across diverse missing-data scenarios.}
    \label{fig:ablationCF}
\end{figure*}
% \begin{figure}[htbp]
%     % \centerline{\includegraphics[width=\columnwidth]{fig/ablationCF.four.pdf}}
%     \centerline{\includegraphics[width=\columnwidth]{fig/ablationCF.pdf}}
%     % \centerline{\includegraphics[width=0.95\columnwidth]{fig/ablation.pdf}}
%     \caption{Ablation study results on the ETTm2 dataset under different missing scenarios. The bars show the relative performance degradation when removing each component, demonstrating their contributions to the overall model performance. Lower values indicate greater importance of the component.}
%     \label{fig:ablationCF1}
% \end{figure}

\subsection{Comparison with State-of-the-art}
Table~\ref{tb:Point} and Table~\ref{tb:Block} report the forecasting results of  CoIFNet and state-of-the-art methods on six real-world datasets at various point and block missing rates, respectively.
As can be seen, our CoIFNet consistently outperforms all competitors under all missing-data scenarios, demonstrating its effectiveness and robustness. Concretely, from the results in Table~\ref{tb:Point}, we see that CoIFNet outperforms the previous state-of-the-art method BiTGraph by \underline{\textbf{18.73}}\% (\underline{\textbf{19.36}}\%) in MAE and \underline{\textbf{22.51}}\% (\underline{\textbf{24.40}}\%) in MSE when the point missing rate is 0.3 (0.6).
% 0.3, MAE: (0.363 - 0.295) / 0.363 * 100 = 18.7327823691
% 0.3, MSE: (0.311 - 0.241) / 0.311 * 100 = 22.5080385852
% 0.6, MAE: (0.377 - 0.304) / 0.377 * 100 = 19.3633952255
% 0.6, MSE: (0.332 - 0.251) / 0.332 * 100 = 24.3975903614
% Interestingly, we find that general MTSF models demonstrate reasonable tolerance to the point missing pattern, with models like iTransformer and PatchTST outperforming BiTGraph, which is specifically designed for missing values scenarios. This suggests that the self-attention mechanisms in transformer-based architectures can effectively capture dependencies even when some data points are randomly missing.
% The block-wise missing scenario (Table~\ref{tb:Block}) presents a substantially more challenging setting, as evidenced by the significant performance degradation observed across all methods compared to the point-wise scenario. 
% 0.3, MAE: (0.386 - 0.305) / 0.386 * 100 = 20.9844559585
% 0.3, MSE: (0.341 - 0.254) / 0.341 * 100 = 25.5131964809
% 0.6, MAE: (0.394 - 0.317) / 0.394 * 100 = 19.5431472081
% 0.6, MSE: (0.357 - 0.272) / 0.357 * 100 = 23.8095238095
From the results in Table~\ref{tb:Block}, we can observe that our CoIFNet outperforms BiTGraph by \underline{\textbf{20.98}}\% (\underline{\textbf{19.54}}\%) in MAE and \underline{\textbf{25.51}}\% (\underline{\textbf{23.81}}\%) in MSE when the block missing rate is 0.3 (0.6).
The superior performance of our CoIFNet over other methods highlight CoIFNet's superior capability in maintaining forecasting accuracy under severe missing conditions.
% In the point missing scenario (Table~\ref{tb:Block}), CoIFNet achieves even more substantial improvements over competing methods, with average performance gains of 25.45\% in MAE and 25.51\% in MSE at missing rate $r=0.3$, and 19.54\% in MAE and 23.81\% in MSE at missing rate $r=0.6$.
Notably, CoIFNet yields lower standard deviations (i.e., $\underline{\pm\textbf{.00}}$) in both MAE and MSE compared to other methods, highlighting its superior stability and robustness.

Additionally, comparing the results of Table~\ref{tb:Point} and Table~\ref{tb:Block}, we have the following observations.
Firstly, time series with missing blocks pose greater challenges than those with missing points for all methods.
This is likely because temporal dependencies are more severely disrupted when consecutive observations are missing.
% This is particularly evident in the Weather and ETT datasets, where the performance degradation from point-wise to block-wise missing is most pronounced.
Secondly, traditional imputation-based approaches like BRITS and SAITS exhibit suboptimal performance in both scenarios, especially at higher missing rates. This can be attributed to their emphasis on reconstruction rather than forecasting, leading to a misalignment between training and evaluation objectives.
% Similarly, linear models like DLinear and STID, while effective for complete time series, lack explicit mechanisms for handling missing values, limiting their robustness in incomplete data scenarios.
% Third, transformer-based architectures (iTransformer, PatchTST) emerge as strong competitors in the point missing pattern but show significant performance drops in the block pattern. One possible reason may be that the self-attention mechanism can work around missing points but struggle when consecutive observations are unavailable. 
Thirdly, transformer-based architectures (i.e., iTransformer, PatchTST) demonstrate competitive performance under the point missing pattern but exhibit significant degradation under the block missing pattern.
This performance disparity likely stems from the inherent limitations of self-attention mechanisms: while they can effectively interpolate missing values through global context aggregation, their ability to interpolate missing blocks of time series is constrained.
Finally, BiTGraph generally shows superior performance than other previous methods, which confirms the importance of explicitly modeling missing patterns when dealing with structured missing values.
However, the superior performance of our CoIFNet over BiTGraph also reveal that BiTGraph's effectiveness is limited by its over-dependence on local spatial correlations and insufficient modeling of cross-timestep and cross-variate interactions.
% Graph-based approaches (MTGNN, AGCRN) face similar challenges as missing values disrupt their graph structure learning.

% The superior performance of CoIFNet across the two missing patterns can be attributed to its unified framework that simultaneously addresses imputation and forecasting tasks. By jointly optimizing for both objectives, our model effectively bridges the gap between historical patterns and future trends, enabling robust forecasting even with limited data availability. The model's consistent performance across different missing rates and patterns highlights the effectiveness of our design in extracting and utilizing available information, demonstrating strong potential for real-world applications where data quality and availability can vary significantly.

\begin{figure*}[htbp]
    \setlength{\abovecaptionskip}{0pt}  % 标题上方间距
    \setlength{\belowcaptionskip}{0pt}  % 标题下方间距
    \centerline{\includegraphics[width=\textwidth]{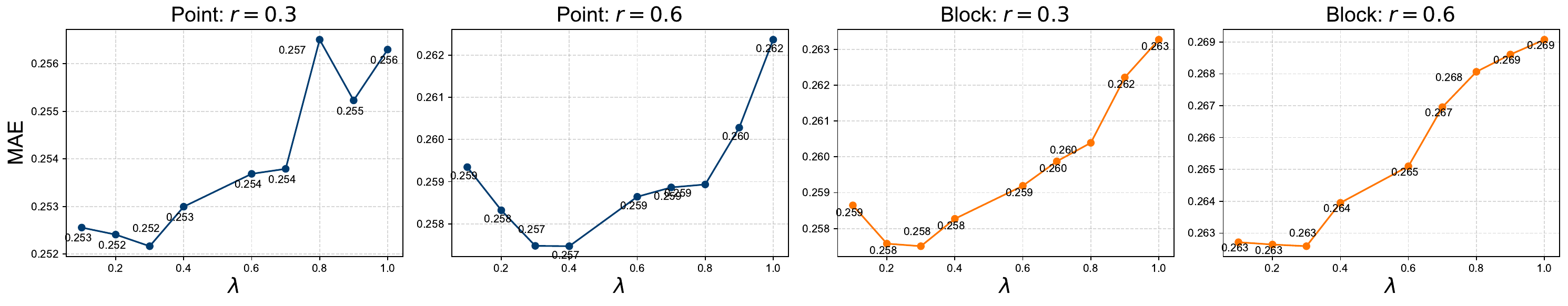}}
    \caption{Ablation study of the balance parameter $\lambda$ on the ETTm2 dataset across diverse missing-data scenarios.}
    \label{fig:model-analysis}
\end{figure*}

\subsection{Ablation Studies}
In this part, we perform ablation studies to scrutinize the effectiveness of the components of CoIFNet and the impact of other design choices on CoIFNet. We conduct experiments on two representative datasets ETTm2 and Weather.
% , we conduct comprehensive ablation studies. These experiments assess the impact of each component on the overall forecasting performance, particularly under varying missing value scenarios.

\noindent{\textbf{Effectiveness of the Designed Components}}:
We investigate the effectiveness of each key component in our proposed CoIFNet by individually removing or replacing elements and measuring the resulting performance.
Table~\ref{tb:ablation} presents the results over the two datasets  under different missing-data scenarios, where “Baseline” indicates that we directly perform vanilla forecasting on incomplete time series data, and “w/o CTF/CVF” denotes the CTF/CVF component is replaced by a linear layer.

% \begin{itemize}
%     % \item \textbf{Baseline vs. CoIFNet}: XXX.
%     \item \textbf{CoIFNet vs. \circled{1}-\circled{7}}: XXX.
%     \item \textbf{\circled{7} vs. \circled{7}-{v1}}: XXX.
% \end{itemize}
% 1.Ett-MAE0.3 (0.263 - 0.252) / 0.263 * 100 = 4.1825095057
% 1.Ett-MAE0.6 (0.271 - 0.258) / 0.271 * 100 = 4.7970479705
% 1.Ett-MSE0.3 (0.189 - 0.172) / 0.189 * 100 = 8.9947089947
% 1.Ett-MSE0.6 (0.200 - 0.179) / 0.200 * 100 = 10.5
% 1.Wea-MAE0.3 (0.212 - 0.200) / 0.212 * 100 = 5.6603773585
% 1.Wea-MAE0.6 (0.218 - 0.204) / 0.218 * 100 = 6.4220183486
% 1.Wea-MSE0.3 (0.169 - 0.157) / 0.169 * 100 = 7.100591716
% 1.Wea-MSE0.6 (0.174 - 0.161) / 0.174 * 100 = 7.4712643678
From the reported results in Table~\ref{tb:ablation}, we have the following observations. Firstly, removing each component of CoIFNet results in performance degradation, demonstrating the effectiveness of these components in enhancing overall performance.
Secondly, from the results of \circled{1}\circled{2}\circled{3}, we observe that incorporating the mask matrix and timestamp embeddings with the input time series data brings performance gains.
Possible reasons are as follows: 1) The mask matrix enables models to distinguish between actual zeros and missing values during training. 2) The timestamp information in time series helps the model capture cyclical patterns, which is critical for improving forecasting performance in the presence of missing data.
% Variants \circled{2}-\circled{4}, which remove different aspects of our fusion mechanism ($\mathbf{M}^x$, $\tau$, and $E^\tau$), all show performance degradation compared to the full CoIFNet model. Specifically, removing the mask matrix fusion ($M^x$) leads to noticeable performance drops, particularly in scenarios with higher missing rates, confirming that explicitly incorporating missing pattern information enhances the model's ability to learn meaningful representations from incomplete data. Secondly, removing timestamp information ($\tau$) (variant \circled{3}) substantially reduces performance, highlighting the importance of proper temporal encoding for capturing time-dependent patterns.
% The combined removal of both mask and timestamp information (variant \circled{4}) further exacerbates performance degradation, demonstrating the complementary nature of these components.
% Thirdly, the most substantial performance drop occurs when removing the imputation task (variant \circled{5}), particularly on the Weather dataset where the reduction in forecasting accuracy is most pronounced.
% This finding is especially significant as Weather represents a dataset with relatively stable patterns and minimal distribution drift. 
Thirdly, from the results of \circled{4}\circled{5}\circled{6}, we see that both the CTF and CVF modules play important roles in performance enhancement. This validates the capability of the two modules to fuse information along the timestep dimension and variate dimension, respectively.
Fourthly, from the results of \circled{7}\circled{8}, we observe a dramatic performance decline across all settings after removing RevON or replacing it with RevIN~\cite{kim2021reversible}.
% The main reason may be that the data drift issue is more pronounced in the ETTm2 dataset compared to the Weather dataset.
% Moreover, replacing our RevON with RevIN~\cite{kim2021reversible} results in a significant decline in CoIFNet's performance across all settings. 
This demonstrates the advantages of RevON in reducing discrepancies among historical sequences in missing-data scenarios.
% Besides, .
The main reason may be that the data drift issue is more pronounced in the ETTm2 dataset compared to the Weather dataset.
Finally, from the results of \circled{9}, we notice that without imputing/reconstructing $\hat{\mathbf{X}}$, the model's performance degrades considerably.
This underscores the pivotal role of the auxiliary imputation task in improving forecasting performance under missing data conditions.

\begin{figure}[t]
    \setlength{\abovecaptionskip}{0pt}  % 标题上方间距
    \setlength{\belowcaptionskip}{0pt}  % 标题下方间距
    % \centerline{\includegraphics[width=0.98\columnwidth]{fig/params.v2.pdf}}
    % \centerline{\includegraphics[width=0.98\columnwidth]{fig/params_broken_axis}}
    \centerline{\includegraphics[width=0.93\columnwidth]{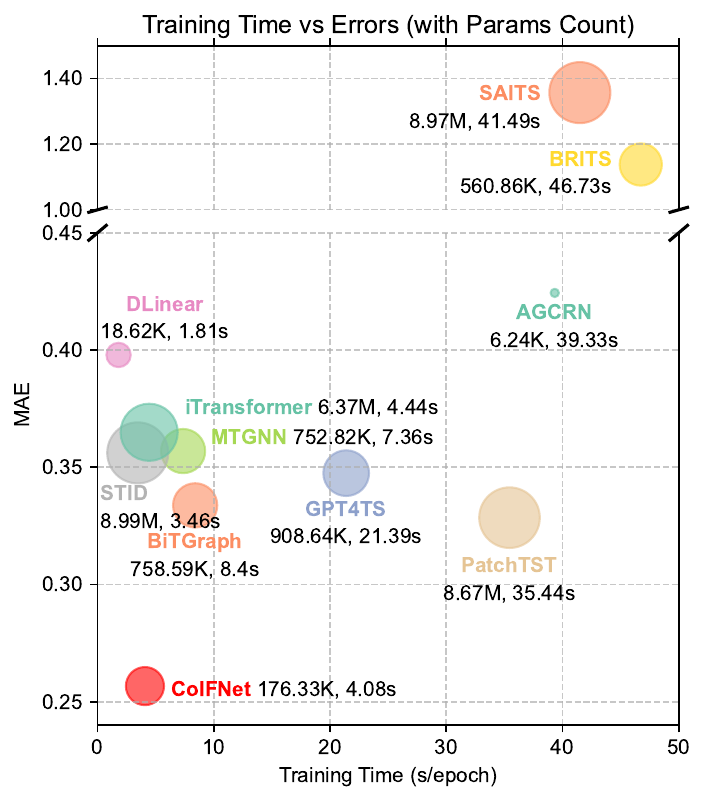}}
    
    \caption{Comparison of forecasting performance and computational efficiency for different methods on the ETTm2 dataset, under the block missing rate of 0.3.}
    %  Each method is represented by a dot, with size scaled according to its params count. Lower values for MSE, training time, and params indicate better performance. Our CoIFNet method demonstrates superior performance across all metrics.
    \label{fig:params}
\end{figure}

\noindent{\textbf{Ablation of Temporal Fusion Components}:
To further investigate the effectiveness of our proposed Cross-Timestep Fusion (CTF) and Cross-Variate Fusion (CVF) modules, we conduct ablation experiments by replacing these components with two alternative networks: Linear and Attention. Linear employs a single-layer MLP for fusion, while Attention utilizes a single-layer multi-head attention mechanism. The obtained results are reported in Fig.~\ref{fig:ablationCF}.
The experimental results reveal several important insights.
Firstly, CTF+CVF achieves the best results compared with other alternatives, demonstrating the effectiveness of our specialized fusion mechanisms  for tackling MSTF in the presence of missing data.
Secondly, variants with CTF, CVF, and Attention consistently outperform that with Linear across all scenarios. This suggests that simple linear transformations are insufficient for effectively modeling the intricate relationships in multivariate time series with missing.
% Thirdly, despite the Attention variant having more parameters, our proposed CTF and CVF modules achieve superior performance. This demonstrates the efficiency and effectiveness of our specialized fusion mechanisms designed specifically for the multivariate time series forecasting task with missing data. % The performance advantage is particularly evident in the block-wise missing scenario, where capturing dependencies across both time steps and variates becomes more challenging due to consecutive missing observations. 
Thirdly, we observe that Attention+CVF achieves comparable results to CTF+CVF in the point missing pattern but it exhibits  significant  performance degradation in the block missing pattern, which highlights the effectiveness and efficiency of our proposed CoIFNet framework.

\noindent\textbf{Impact of the Balance Weight $\lambda$.}
In the proposed CoIFNet, we employ the balance weight $\lambda$ to control the relative importance of the imputation and forecasting losses.
It is necessary to investigate the impact of $\lambda$ on the performance of CoIFNet.
To this end, we set $\lambda$ to the values of $\{0.1, 0.2, ..., 1.0\}$, and report the average testing results on the 6 datasets in Fig.~\ref{fig:model-analysis}.
Overall, the performance of CoIFNet gradually increases as the $\lambda$ value grows from 0.1 to 0.3, after which the performance of DePT gradually decreases and reaches the lowest value when $\lambda=1.0$.
In particular, when $\lambda=0.3$ CoIFNet establishes the best performance across all those settings.
% We also see that when $\lambda$ takes the values of $\{0.1, ..., 0.9\}$, DePT consistently outperforms the baseline (i.e., when $\lambda$=0.0), demonstrating the effectiveness of the designed CAT head and our DePT framework.
What is noteworthy is that when $\lambda=1.0$, i.e., only the forecasting loss is used for training, the performance of CoIFNet sharply decreases, which suggests that joint optimization of imputation and forecasting objectives is of great importance for achieving better MTSF performance in the presence of missing values.

\subsection{Computational Efficiency Analysis}\label{exp:comput}

\begin{figure*}[htbp]
    \setlength{\abovecaptionskip}{0.5pt}  % 标题上方间距
    \setlength{\belowcaptionskip}{0pt}  % 标题下方间距
    \centerline{\includegraphics[width=0.97\textwidth]{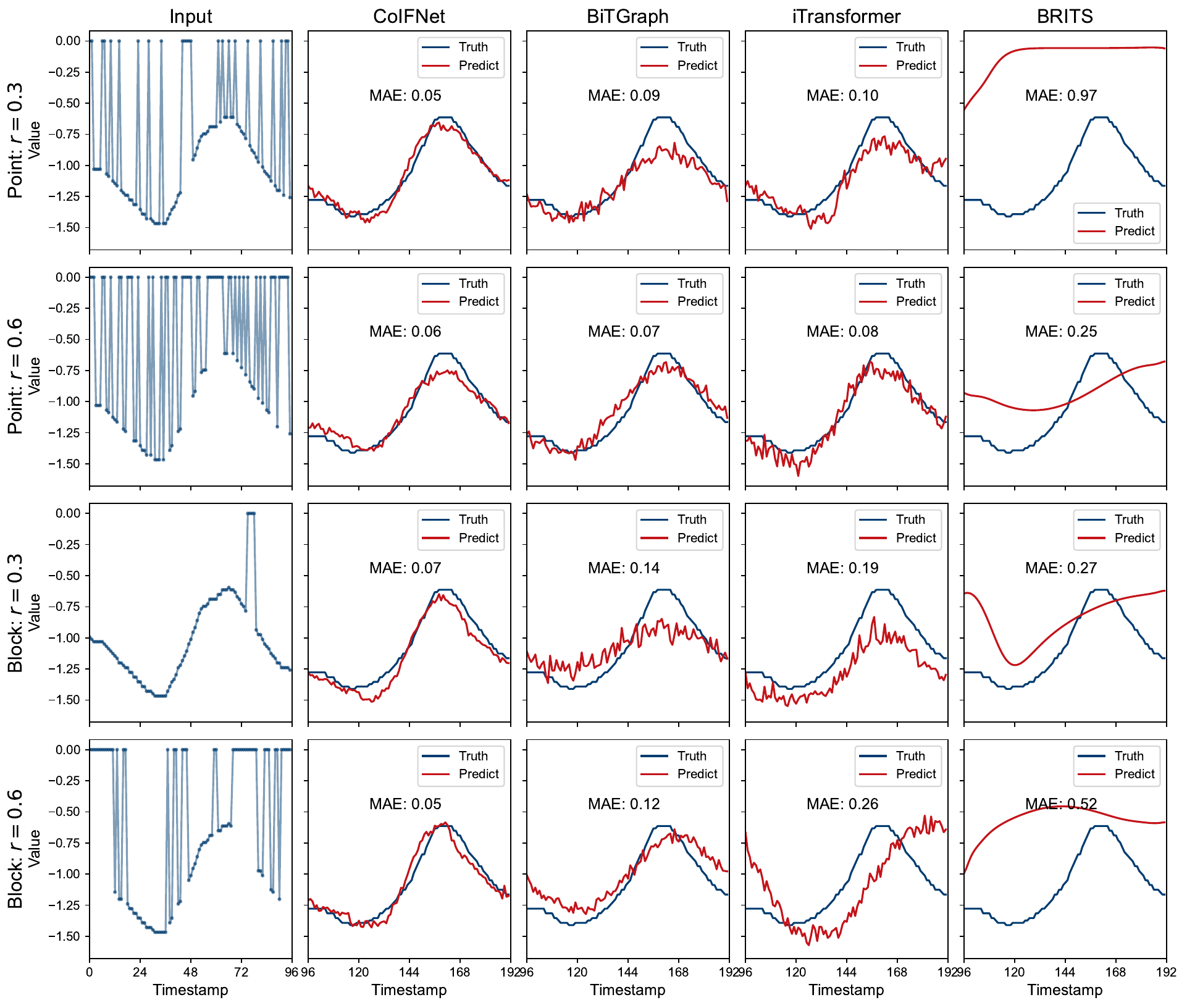}}
    \caption{Forecasting curves of BiTGraph, iTransformer, BRITS and our CoIFNet on the ETTm2 dataset.}
    \label{fig:case}
\end{figure*}

Beyond forecasting accuracy, computational efficiency plays a crucial role in practical applications of time series models. We conduct a comprehensive efficiency analysis comparing CoIFNet with mainstream baselines in terms of model parameters and training time. The results are summarized in Fig.~\ref{fig:params}.
As demonstrated, the efficiency advantages of CoIFNet manifest through three key aspects. Firstly,
% compared to BiTGraph, the previous state-of-the-art model for MTSF with missing values, CoIFNet demonstrates remarkable efficiency with only 24.05\% (758.59K$\rightarrow$182.47K) of the parameters and 25.18\% (32.49M $\rightarrow$8.18M) of the MACs while achieving better performance. 
compared to BiTGraph, the previous state-of-the-art model for MTSF with missing values, CoIFNet achieves a $\underline{\boldsymbol{4.3\times}}$ improvement in memory efficiency and a  $\underline{\boldsymbol{2.1\times}}$  gain in time efficiency.
This efficiency stems from our carefully designed CTF and CVF module, which captures temporal dependencies through lightweight linear operations rather than quadratic-complexity attention mechanisms.
% This can be attributed to our lightweight architecture for modeling temporal dependencies.
Secondly, CoIFNet demonstrates significant computational efficiency advantages over Transformer-based architectures such as PatchTST, iTransformer, and GPT4TS. Although these models achieve improved predictive performance by leveraging self-attention mechanisms, their reliance on such operations results in substantial computational overhead.
% Our model maintains comparable or superior accuracy with approximately 80\%-97\% fewer parameters and 10\%-99\% fewer MACs.
Thirdly, even compared to Linear-based models like STID, CoIFNet demonstrates comparable or superior computational efficiency.
The main reason is that these rivals require more computational resources due to their relying on complex temporal dependence modeling modules.
% Our CoIFNet achieves superior results with approximately 99\% fewer parameters through its efficient integration of temporal features and missing patterns.
The lightweight architecture enables faster training and inference while maintaining robust performance across various missing patterns and missing rates. Those advantages render CoIFNet especially suitable for real-world deployments where both computational efficiency and forecasting accuracy are critical considerations.

\subsection{Case Study}
To provide deeper insights into model behavior, we conduct detailed case studies on the ETTm2 dataset under different missing scenarios. Fig.~\ref{fig:case} presents the forecasting results of our proposed CoIFNet and three representative paradigms: BiTGraph, iTransformer, and BRITS.
% We examine four challenging scenarios: point-wise missing with 30\% and 60\% rates, and block-wise missing with 30\% and 60\% rates in Fig.~\ref{fig:case}.
At a point missing rate  of 0.3, we observe that while iTransformer captures the general trend of the time series, it struggles to accurately predict the magnitude of peaks and troughs. BiTGraph demonstrates improved performance over iTransformer, confirming the value of explicitly modeling missing patterns. However, CoIFNet achieves the closest alignment with the ground truth, particularly in capturing the amplitude variations and temporal dynamics. BRITS, despite being designed for handling missing values, shows significant deviations from the ground truth, indicating that imputation-focused approaches without explicit forecasting optimization fail to adapt well to the forecasting task. As the missing rate increases to 0.6, the performance gap between different models widens. BRITS exhibits substantial prediction errors, iTransformer maintains reasonable trend prediction but misses important fluctuations in the time series. BiTGraph performs better than iTransformer, demonstrating the importance of specialized modeling for missing values scenarios. CoIFNet continues to provide the most accurate forecasts, with predictions that closely follow the ground truth curve even under this challenging high-missing-rate condition.
At a block missing rate of 0.3, BRITS and iTransformer struggle significantly, failing to capture the temporal dynamics after extended missing segments. BiTGraph shows improved performance but still exhibits noticeable deviations from the ground truth. CoIFNet maintains superior prediction accuracy, effectively reconstructing the underlying patterns despite the presence of large missing blocks in the input data.
As the missing rate increases to 0.6, the performance difference becomes most pronounced. BRITS produces forecasts that bear little resemblance to the actual time series, while iTransformer captures only the most basic trend. BiTGraph manages to approximate some features of the time series but misses critical temporal transitions. In contrast, CoIFNet continues to generate forecasts that closely align with the ground truth, demonstrating CoIFNet's robustness against missing values.

% These visualizations empirically validate the effectiveness of our collaborative imputation-forecasting approach in handling diverse missing patterns. The superior performance of CoIFNet is particularly evident in challenging scenarios with high missing rates or extended missing blocks, where traditional methods often fail to maintain prediction stability. By jointly optimizing for both imputation and forecasting objectives, CoIFNet develops more robust representations of the underlying temporal dynamics, enabling accurate predictions even with severely limited input information.

\subsection{Additional Results} \label{app.0.1}
Table~\ref{tb:add} compares the time series forecasting performance of our CoIFNet approach with state-of-the-art methods on six real-world datasets, at a point/block missing rate of 0.1.
As can be observed, CoIFNet achieves the best or comparable results across the six datasets demonstrating its effectiveness and robustness. On average, CoIFNet outperforms the previous state-of-the-art method BiTGraph by \underline{\textbf{15.56}}\% (\underline{\textbf{17.36}}\%) in MAE and \underline{\textbf{17.96}}\% (\underline{\textbf{21.36}}\%) in MSE when the point (block) missing rate is 0.1.

\begin{table*}[!t]
    \caption{Comparison with state-of-the-arts on six real-world datasets under the \textbf{{Point}}/\textbf{{Block}} missing rate  of 0.1.}
    \centering
    \setlength{\tabcolsep}{3pt}
    \resizebox{\textwidth}{!}{
        \begin{tabular}{c|cc|cc|cc|cc|cc|cc|cc}
            \toprule
            \multicolumn{15}{c}{\textbf{Point}, $r=0.1$}                                                                                                                                                                                                                                                                                                                                                                                                                        \\
            \hline
            Dataset                              & \multicolumn{2}{c|}{ETTh1} & \multicolumn{2}{c|}{ETTh2} & \multicolumn{2}{c|}{ETTm1} & \multicolumn{2}{c|}{ETTm2} & \multicolumn{2}{c|}{Weather} & \multicolumn{2}{c|}{Exchange Rate} & \multicolumn{2}{c}{\textbf{AVG}}                                                                                                                                                                                                     \\
            Metric                               & MAE                        & MSE                        & MAE                        & MSE                        & MAE                          & MSE                                & MAE                              & MSE                       & MAE                       & MSE                       & MAE                       & MSE                       & MAE                       & MSE                       \\
            \hline
            BRITS                                & 0.811$_{\pm.01}$           & 1.203$_{\pm.02}$           & 1.169$_{\pm.04}$           & 2.364$_{\pm.08}$           & 0.779$_{\pm.03}$             & 1.070$_{\pm.02}$                   & 1.168$_{\pm.11}$                 & 2.276$_{\pm.49}$          & 0.476$_{\pm.07}$          & 0.472$_{\pm.09}$          & 1.367$_{\pm.03}$          & 2.776$_{\pm.10}$          & 0.962$_{\pm.05}$          & 1.693$_{\pm.13}$          \\
            SAITS                                & 0.841$_{\pm.00}$           & 1.268$_{\pm.00}$           & 1.187$_{\pm.01}$           & 2.437$_{\pm.03}$           & 0.789$_{\pm.00}$             & 1.097$_{\pm.01}$                   & 1.353$_{\pm.00}$                 & 3.108$_{\pm.02}$          & 0.587$_{\pm.00}$          & 0.611$_{\pm.00}$          & 1.397$_{\pm.04}$          & 2.916$_{\pm.13}$          & 1.026$_{\pm.01}$          & 1.906$_{\pm.03}$          \\
            % \hline
            STID                                 & 0.450$_{\pm.00}$           & 0.475$_{\pm.00}$           & 0.338$_{\pm.00}$           & 0.258$_{\pm.00}$           & 0.360$_{\pm.00}$             & 0.330$_{\pm.00}$                   & 0.280$_{\pm.00}$                 & 0.193$_{\pm.00}$          & 0.204$_{\pm.00}$          & 0.158$_{\pm.00}$          & 0.262$_{\pm.02}$          & 0.128$_{\pm.02}$          & 0.316$_{\pm.00}$          & 0.257$_{\pm.00}$          \\
            MTGNN                                & 0.491$_{\pm.01}$           & 0.514$_{\pm.01}$           & 0.438$_{\pm.01}$           & 0.395$_{\pm.04}$           & 0.406$_{\pm.01}$             & 0.385$_{\pm.01}$                   & 0.319$_{\pm.01}$                 & 0.228$_{\pm.01}$          & 0.222$_{\pm.00}$          & 0.174$_{\pm.00}$          & 0.315$_{\pm.01}$          & 0.178$_{\pm.01}$          & 0.365$_{\pm.01}$          & 0.312$_{\pm.02}$          \\
            AGCRN                                & 0.660$_{\pm.03}$           & 0.851$_{\pm.04}$           & 0.592$_{\pm.01}$           & 0.770$_{\pm.04}$           & 0.492$_{\pm.01}$             & 0.498$_{\pm.02}$                   & 0.350$_{\pm.01}$                 & 0.277$_{\pm.01}$          & 0.218$_{\pm.01}$          & 0.171$_{\pm.01}$          & 0.424$_{\pm.03}$          & 0.325$_{\pm.05}$          & 0.456$_{\pm.02}$          & 0.482$_{\pm.03}$          \\
            DLinear                              & 0.445$_{\pm.00}$           & \textbf{0.469$_{\pm.00}$}  & 0.338$_{\pm.00}$           & 0.256$_{\pm.00}$           & 0.365$_{\pm.00}$             & 0.346$_{\pm.00}$                   & 0.288$_{\pm.00}$                 & 0.203$_{\pm.00}$          & 0.247$_{\pm.00}$          & 0.207$_{\pm.00}$          & 0.233$_{\pm.00}$          & 0.101$_{\pm.00}$          & 0.319$_{\pm.00}$          & 0.264$_{\pm.00}$          \\
            PatchTST                             & 0.454$_{\pm.00}$           & 0.476$_{\pm.00}$           & 0.333$_{\pm.00}$           & 0.257$_{\pm.00}$           & 0.363$_{\pm.00}$             & 0.342$_{\pm.00}$                   & 0.276$_{\pm.00}$                 & 0.196$_{\pm.00}$          & 0.222$_{\pm.00}$          & 0.181$_{\pm.00}$          & 0.229$_{\pm.01}$          & 0.100$_{\pm.01}$          & 0.313$_{\pm.00}$          & 0.259$_{\pm.00}$          \\
            iTransformer                         & 0.470$_{\pm.00}$           & 0.488$_{\pm.01}$           & 0.368$_{\pm.00}$           & 0.296$_{\pm.01}$           & 0.384$_{\pm.00}$             & 0.349$_{\pm.00}$                   & 0.301$_{\pm.00}$                 & 0.218$_{\pm.00}$          & 0.236$_{\pm.00}$          & 0.183$_{\pm.00}$          & 0.287$_{\pm.01}$          & 0.163$_{\pm.01}$          & 0.341$_{\pm.00}$          & 0.283$_{\pm.00}$          \\
            GPT4TS                               & 0.448$_{\pm.00}$           & 0.475$_{\pm.01}$           & 0.351$_{\pm.01}$           & 0.274$_{\pm.01}$           & 0.362$_{\pm.00}$             & 0.340$_{\pm.00}$                   & 0.278$_{\pm.00}$                 & 0.197$_{\pm.00}$          & 0.228$_{\pm.00}$          & 0.188$_{\pm.00}$          & 0.234$_{\pm.01}$          & 0.104$_{\pm.00}$          & 0.317$_{\pm.00}$          & 0.263$_{\pm.00}$          \\
            % \hline
            BiTGraph                             & 0.485$_{\pm.00}$           & 0.507$_{\pm.00}$           & 0.386$_{\pm.01}$           & 0.313$_{\pm.01}$           & 0.404$_{\pm.01}$             & 0.381$_{\pm.00}$                   & 0.289$_{\pm.01}$                 & 0.194$_{\pm.01}$          & 0.209$_{\pm.00}$          & 0.163$_{\pm.00}$          & 0.311$_{\pm.01}$          & 0.171$_{\pm.01}$          & 0.347$_{\pm.01}$          & 0.288$_{\pm.01}$          \\
            \hline
            \rowcolor{green!10} \textbf{CoIFNet} & \textbf{0.445$_{\pm.00}$}  & 0.471$_{\pm.00}$           & \textbf{0.313$_{\pm.00}$}  & \textbf{0.231$_{\pm.00}$}  & \textbf{0.346$_{\pm.00}$}    & \textbf{0.313$_{\pm.00}$}          & \textbf{0.249$_{\pm.00}$}        & \textbf{0.169$_{\pm.00}$} & \textbf{0.198$_{\pm.00}$} & \textbf{0.156$_{\pm.00}$} & \textbf{0.206$_{\pm.00}$} & \textbf{0.088$_{\pm.00}$} & \textbf{0.293$_{\pm.00}$} & \textbf{0.238$_{\pm.00}$} \\
            \hline \hline

            \multicolumn{15}{c}{\textbf{Block}, $r=0.1$}                                                                                                                                                                                                                                                                                                                                                                                                                        \\
            \hline

            \hline
            Dataset                              & \multicolumn{2}{c|}{ETTh1} & \multicolumn{2}{c|}{ETTh2} & \multicolumn{2}{c|}{ETTm1} & \multicolumn{2}{c|}{ETTm2} & \multicolumn{2}{c|}{Weather} & \multicolumn{2}{c|}{Exchange Rate} & \multicolumn{2}{c}{\textbf{AVG}}                                                                                                                                                                                                     \\
            Metric                               & MAE                        & MSE                        & MAE                        & MSE                        & MAE                          & MSE                                & MAE                              & MSE                       & MAE                       & MSE                       & MAE                       & MSE                       & MAE                       & MSE                       \\
            \hline
            BRITS                                & 0.802$_{\pm.01}$           & 1.185$_{\pm.03}$           & 1.076$_{\pm.10}$           & 2.025$_{\pm.41}$           & 0.780$_{\pm.02}$             & 1.067$_{\pm.03}$                   & 1.246$_{\pm.05}$                 & 2.632$_{\pm.26}$          & 0.426$_{\pm.01}$          & 0.400$_{\pm.01}$          & 1.235$_{\pm.11}$          & 2.202$_{\pm.36}$          & 0.928$_{\pm.05}$          & 1.585$_{\pm.18}$          \\
            SAITS                                & 0.847$_{\pm.00}$           & 1.275$_{\pm.00}$           & 1.194$_{\pm.00}$           & 2.484$_{\pm.01}$           & 0.792$_{\pm.00}$             & 1.105$_{\pm.00}$                   & 1.352$_{\pm.01}$                 & 3.112$_{\pm.03}$          & 0.589$_{\pm.00}$          & 0.613$_{\pm.00}$          & 1.397$_{\pm.04}$          & 2.919$_{\pm.14}$          & 1.029$_{\pm.01}$          & 1.918$_{\pm.03}$          \\
            % \hline
            STID                                 & 0.462$_{\pm.00}$           & 0.485$_{\pm.01}$           & 0.368$_{\pm.01}$           & 0.310$_{\pm.01}$           & 0.373$_{\pm.01}$             & 0.347$_{\pm.00}$                   & 0.310$_{\pm.00}$                 & 0.243$_{\pm.01}$          & 0.207$_{\pm.00}$          & 0.162$_{\pm.00}$          & 0.280$_{\pm.01}$          & 0.161$_{\pm.01}$          & 0.334$_{\pm.01}$          & 0.285$_{\pm.01}$          \\
            MTGNN                                & 0.497$_{\pm.00}$           & 0.518$_{\pm.00}$           & 0.440$_{\pm.03}$           & 0.415$_{\pm.05}$           & 0.419$_{\pm.00}$             & 0.407$_{\pm.01}$                   & 0.336$_{\pm.01}$                 & 0.245$_{\pm.00}$          & 0.219$_{\pm.00}$          & 0.171$_{\pm.00}$          & 0.354$_{\pm.01}$          & 0.238$_{\pm.02}$          & 0.378$_{\pm.01}$          & 0.332$_{\pm.01}$          \\
            AGCRN                                & 0.648$_{\pm.01}$           & 0.826$_{\pm.01}$           & 0.635$_{\pm.03}$           & 0.876$_{\pm.07}$           & 0.509$_{\pm.02}$             & 0.522$_{\pm.03}$                   & 0.430$_{\pm.05}$                 & 0.438$_{\pm.12}$          & 0.222$_{\pm.00}$          & 0.174$_{\pm.00}$          & 0.473$_{\pm.02}$          & 0.396$_{\pm.04}$          & 0.486$_{\pm.02}$          & 0.539$_{\pm.05}$          \\
            DLinear                              & \textbf{0.450$_{\pm.00}$}  & \textbf{0.478$_{\pm.00}$}  & 0.371$_{\pm.00}$           & 0.314$_{\pm.00}$           & 0.381$_{\pm.00}$             & 0.372$_{\pm.00}$                   & 0.331$_{\pm.01}$                 & 0.278$_{\pm.01}$          & 0.257$_{\pm.00}$          & 0.218$_{\pm.00}$          & 0.285$_{\pm.01}$          & 0.170$_{\pm.01}$          & 0.346$_{\pm.00}$          & 0.305$_{\pm.00}$          \\
            PatchTST                             & 0.459$_{\pm.00}$           & 0.486$_{\pm.00}$           & 0.349$_{\pm.00}$           & 0.280$_{\pm.01}$           & 0.378$_{\pm.00}$             & 0.367$_{\pm.00}$                   & 0.289$_{\pm.00}$                 & 0.220$_{\pm.00}$          & 0.233$_{\pm.00}$          & 0.192$_{\pm.00}$          & 0.250$_{\pm.02}$          & 0.271$_{\pm.24}$          & 0.326$_{\pm.01}$          & 0.303$_{\pm.04}$          \\
            iTransformer                         & 0.474$_{\pm.00}$           & 0.494$_{\pm.00}$           & 0.411$_{\pm.00}$           & 0.372$_{\pm.00}$           & 0.404$_{\pm.01}$             & 0.386$_{\pm.02}$                   & 0.348$_{\pm.01}$                 & 0.281$_{\pm.02}$          & 0.242$_{\pm.00}$          & 0.187$_{\pm.00}$          & 0.327$_{\pm.00}$          & 0.228$_{\pm.01}$          & 0.368$_{\pm.00}$          & 0.325$_{\pm.01}$          \\
            GPT4TS                               & 0.451$_{\pm.00}$           & 0.482$_{\pm.00}$           & 0.368$_{\pm.00}$           & 0.314$_{\pm.01}$           & 0.372$_{\pm.00}$             & 0.361$_{\pm.00}$                   & 0.297$_{\pm.01}$                 & 0.234$_{\pm.01}$          & 0.231$_{\pm.00}$          & 0.192$_{\pm.00}$          & 0.252$_{\pm.00}$          & 0.136$_{\pm.01}$          & 0.329$_{\pm.00}$          & 0.286$_{\pm.00}$          \\
            % \hline
            BiTGraph                             & 0.498$_{\pm.01}$           & 0.520$_{\pm.01}$           & 0.414$_{\pm.04}$           & 0.359$_{\pm.05}$           & 0.413$_{\pm.01}$             & 0.394$_{\pm.00}$                   & 0.312$_{\pm.01}$                 & 0.217$_{\pm.01}$          & 0.215$_{\pm.00}$          & 0.167$_{\pm.00}$          & 0.323$_{\pm.01}$          & 0.199$_{\pm.01}$          & 0.362$_{\pm.01}$          & 0.309$_{\pm.02}$          \\
            \hline
            \rowcolor{green!10} \textbf{CoIFNet} & 0.452$_{\pm.00}$           & 0.480$_{\pm.00}$           & \textbf{0.315$_{\pm.00}$}  & \textbf{0.235$_{\pm.00}$}  & \textbf{0.353$_{\pm.00}$}    & \textbf{0.323$_{\pm.00}$}          & \textbf{0.252$_{\pm.00}$}        & \textbf{0.173$_{\pm.00}$} & \textbf{0.200$_{\pm.00}$} & \textbf{0.158$_{\pm.00}$} & \textbf{0.209$_{\pm.00}$} & \textbf{0.089$_{\pm.00}$} & \textbf{0.297$_{\pm.00}$} & \textbf{0.243$_{\pm.00}$} \\

            \bottomrule
        \end{tabular}
    }
    \label{tb:add}
\end{table*}

\section{Conclusion}

In this work, we propose CoIFNet, a unified and end-to-end framework for achieving robust multivariate time series forecasting (MTSF) in the presence of missing values. CoIFNet integrates the observed values, mask matrix, and timestamp embeddings, and processes them through sequential Cross-Timestep Fusion (CTF) and Cross-Variate Fusion (CVF) modules to capture temporal dependencies resilient to incomplete observations.
We further provide theoretical justifications based on mutual information, proving the superiority of our one-stage CoIFNet framework over traditional two-stage approaches for handling missing values in time series forecasting.
Extensive experiments on six real-world datasets demonstrate the effectiveness and computational efficiency of our proposed approach.
We hope the proposed CoIFNet framework and its theoretical justifications could inspire future research in related fields.
Our code is  available at: \textcolor{magenta}{\url{https://github.com/KaiTang-eng/CoIFNet}}.

\appendices

\ifCLASSOPTIONcaptionsoff
    \newpage
\fi

\bibliographystyle{IEEEtran}
\bibliography{ref}

\end{document}